\pdfoutput=1
\documentclass[acmsmall,dvipsnames, svgnames, x11names, natbib=true]{acmart}

\usepackage{tikz}
\usepackage{xcolor}
\usetikzlibrary{arrows}
\usepackage{graphicx}
\usepackage{subfigure}
\usetikzlibrary{shapes}
\usepackage{pgfplots}

\usepackage{amsmath}
\usepackage{amsthm}
\usepackage{booktabs}
\usepackage{algorithm}
\usepackage{algorithmic}
\usepackage{multirow}
\usepackage{amsfonts} %
\def\modelname{SHA-LRT}

\usepackage[misc]{ifsym}
\usepackage{bbding}
\usepackage{balance}

\AtBeginDocument{%
  }

\setcopyright{acmcopyright}
\copyrightyear{2018}
\acmYear{2018}
\acmDOI{XXXXXXX.XXXXXXX}

\acmJournal{JACM}
\acmVolume{37}
\acmNumber{4}
\acmArticle{111}
\acmMonth{8}

\begin{document}

\title{Capture Salient Historical Information: A Fast and Accurate Non-Autoregressive Model for Multi-turn Spoken Language Understanding}

\author{Lizhi Cheng}
\email{clz19960630@sjtu.edu.cn}
\affiliation{%
  \institution{Shanghai Jiao Tong University}
  \city{Shanghai}
  \country{PR China}
}

\author{Weijia Jia}
\authornotemark[1]
\email{jiawj@bnu.edu.cn}
\affiliation{%
  \institution{
    BNU-UIC Institute of Artificial Intelligence and Future Networks, Beijing Normal University (Zhuhai),
    Guangdong Key Lab of AI and Multi-Modal Data Processing,
    BNU-HKBU United International College}
  \city{Zhuhai}
  \state{Guangdong}
  \country{PR China}
}

\author{Wenmian Yang}
\authornotemark[1]
\email{wenmian.yang@ntu.edu.sg}
\affiliation{
  \institution{Nanyang Technological University}
  \city{Singapore}
  \country{Singapore}
}

\thanks{*corresponding authors}
\renewcommand{\shortauthors}{Trovato et al.}

\begin{abstract}
Spoken Language Understanding (SLU), a core component of the task-oriented dialogue system, expects a shorter inference facing the impatience of human users. Existing work increases inference speed by designing non-autoregressive models for single-turn SLU tasks but fails to apply to multi-turn SLU in confronting the dialogue history. The intuitive idea is to concatenate all historical utterances and utilize the non-autoregressive models directly. However, this approach seriously misses the salient historical information and suffers from the uncoordinated-slot problems. To overcome those shortcomings, we propose a novel model for multi-turn SLU named Salient History Attention with Layer-Refined Transformer (SHA-LRT), which composes of an SHA module, a Layer-Refined Mechanism (LRM), and a Slot Label Generation (SLG) task. SHA captures salient historical information for the current dialogue from both historical utterances and results via a well-designed history-attention mechanism. LRM predicts preliminary SLU results from Transformer's middle states and utilizes them to guide the final prediction, and SLG obtains the sequential dependency information for the non-autoregressive encoder. Experiments on public datasets indicate that our model significantly improves multi-turn SLU performance (17.5\% on Overall) with accelerating (nearly 15 times) the inference process over the state-of-the-art baseline as well as effective on the single-turn SLU tasks.
\end{abstract}



\maketitle

\section{Introduction}
\begin{figure}[h]
\centering
\tikzstyle{background1} = [rectangle, rounded corners=1mm,minimum width = 13.5cm, minimum height = 3.5cm, text centered, draw = black]
\tikzstyle{background2} = [rectangle, rounded corners=1mm,minimum width = 13.5cm, minimum height = 3.5cm, text centered, draw = black]

\tikzstyle{word} = [rectangle,rounded corners=0.5mm,minimum width = 1.3cm, minimum height = 0.75cm, text centered, thick]
\tikzstyle{slot} = [rectangle,rounded corners=0.5mm,minimum width = 1.3cm, minimum height = 0.75cm, text centered, thick]

\begin{tikzpicture}[node distance = 0cm]

\tikzstyle{every node}=[scale=0.8]
\node(word1)[word,xshift=0cm,yshift=4.5cm]{What};
\node(word1_2)[word,right of=word1,xshift=1cm,yshift=0]{is};
\node(word2)[word,right of=word1_2,xshift=1cm,yshift=0]{the};
\node(word3)[word,right of=word2,xshift=1cm,yshift=0]{address};
\node(word4)[word,right of=word3,xshift=1cm,yshift=0]{for};
\node(word5)[word,right of=word4,xshift=1cm,yshift=0]{Hotel};
\node(word6)[word,right of=word5,xshift=1cm,yshift=0]{A};
\node(Utterance)[left of=word1,xshift=-1.5cm,yshift=0cm]{\textbf{Utterance}:};

\node(slot1)[slot,below of=word1,xshift=0cm,yshift=-1.1cm]{O};
\node(slot1_2)[slot,right of=slot1,xshift=1cm,yshift=0]{O};
\node(slot2)[slot,right of=slot1_2,xshift=1cm,yshift=0]{O};
\node(slot3)[slot,right of=slot2,xshift=1cm,yshift=0]{O};
\node(slot4)[slot,right of=slot3,xshift=1cm,yshift=0]{O};
\node(slot5)[slot,right of=slot4,xshift=1cm,yshift=0]{B-poi};
\node(slot6)[slot,right of=slot5,xshift=1cm,yshift=0]{I-poi};
\node(slots)[left of=slot1,xshift=-1.8cm,yshift=0cm ]{\textbf{Slots}:};
\node(Intent)[below of=slots,xshift=0.075cm,yshift=-0.7cm ]{\textbf{Intent}:};
\node()[below of=slot1,xshift=0cm,yshift=-0.7cm ]{Navigate};

\node(bg1)[background1,below of=slot2,xshift=0.1cm,yshift=0.3cm ]{};
\node(ct)[below of=bg1,xshift=0.0cm,yshift=1.3cm]{\textbf{Current Turn}};

\node()[below of=bg1,xshift=0cm,yshift=7.5cm,align=left]{\textbf{Dialogue History}};
\node(t1)[background2,below of=bg1,xshift=0cm,yshift=5cm,align=left]{
\\
\\
\textbf{Turn 1}:\\
\ \ \textbf{Utterance}: Is there a nearby rest stop ?\\
\ \ \textbf{Slots}: O O O B-distance B-poi\_type I-poi\_type O\\
\ \ \textbf{Intent}: Navigate\\
\ \ \textbf{AI assistant response}: Hotel A and Hotel B are both nearby.\\
\\
\textbf{Turn 2}:\\
\ \ \textbf{Utterance}: Which has the least amount of traffic at the moment ?\\
\ \ \textbf{Slots}: O O O B-traffic\_info I-traffic\_info I-traffic\_info I-traffic\_info B-time I-time I-time O\\
\ \ \textbf{Intent}: Navigate\\
\ \ \textbf{AI assistant response}: They both have none.\\
};

\draw[-latex,thin] (word1.south) -- (slot1.north); 
\draw[-latex,thin] (word1_2.south) -- (slot1_2.north);
\draw[-latex,thin] (word2.south) -- (slot2.north); 
\draw[-latex,thin] (word3.south) -- (slot3.north);
\draw[-latex,thin] (word4.south) -- (slot4.north); 
\draw[-latex,thin] (word5.south) -- (slot5.north); 
\draw[-latex,thin] (word6.south) -- (slot6.north); 
\end{tikzpicture}
\caption{An example of Multi-turn SLU. This example is a real case from KVRET dataset.}
\label{fig:example}
\end{figure}
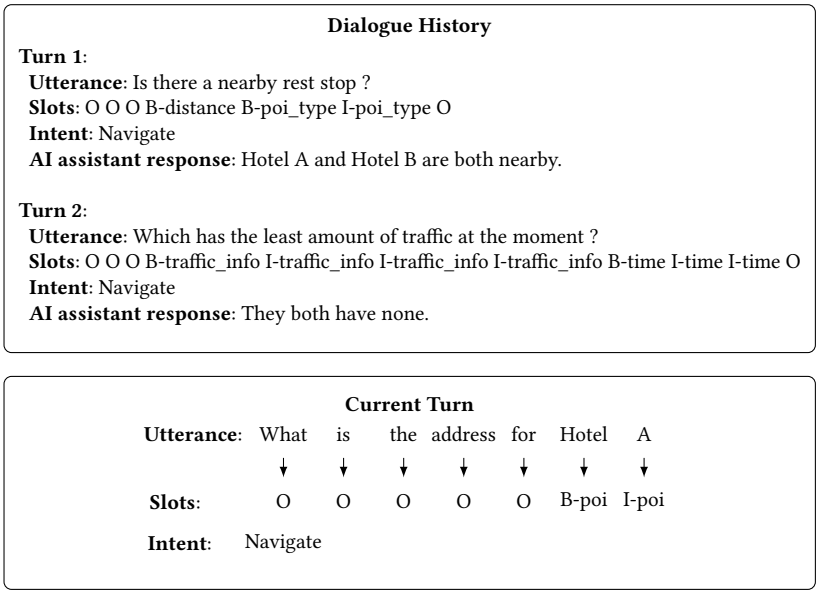

Task-oriented dialogue systems have received more attention with the widespread application of intelligent voice assistants, e.g., Apple Siri, Microsoft Cortana, etc. Working as the spoken interface between users and machines, Spoken Language Understanding (SLU) plays a critical role in the task-oriented dialogue system \cite{TOIS2019Memory,TOIS2021Response,TOIS2021ALargescale}. A typical SLU task mainly includes two subtasks \cite{SLU2011}, i.e., Intent Detection (ID) and Slot Filling (SF). Given by an utterance expressed in natural language from the user, ID aims to identify the intent of the user (e.g., Navigate), and SF aims to fill the slot for each token in the utterance (e.g., distance, time). Generally, ID works on sentence-level and is treated as a semantic classification task, while SF is a sequence labeling task focusing on token-level. 
In recent real application scenarios, people are not satisfied with only making single-turn dialogues with the voice assistant and desire a multi-turn dialogue to achieve their demand. Therefore, the multi-turn dialogue-based SLU task is increasingly crucial \cite{TOISgao2021learning,TOISLi2021Dialogue,TOISVuong2021}. 
A simple example of multi-turn SLU is shown in Fig. \ref{fig:example}.

In multi-turn dialogue, the current utterance is closely related to the dialogue history due to the conversational continuity. Thus, SLU models have to deal with both the dialogue history and the current utterance to enhance the performance, which leads to a large amount of extra computation. Due to the above reason, reducing the inference time of multi-turn SLU tasks is entirely meaningful. However, most existing models process dialogue history inefficiently. Some studies \cite{End-to-end2016,sequentialSlu2017,memory2019} obtain historical information via some lengthy and complicated architectures and thus consume a lot of inference time. \citet{clzICME2021} propose RPFSLU framework which employ an RNN-based portable plugin to obtain historical information for state-of-art single-turn SLU models \cite{bi-dictional2019, qin2019stack}. 
Unfortunately, RPFSLU still needs to calculate the hidden states of each historical utterance in succession, which is low efficiency. Besides, most of the work relies on the RNN-based framework, which models the entire sequence dependency in an autoregressive way, leading to redundant computation and inevitable high latency.

\begin{figure}[h]
\centering
\tikzstyle{background1} = [rectangle, rounded corners=3mm,minimum width = 14cm, minimum height = 5cm, text centered, draw = black, fill = gray!8]
\tikzstyle{memory_list1} = [rectangle,rounded corners=1mm,minimum width = 4cm, minimum height = 0.6cm, text centered, draw = Peru, fill =Peach!15,thick]
\tikzstyle{memory_list2} = [rectangle,rounded corners=1mm,minimum width = 4cm, minimum height = 1.5cm, text centered, draw = Peru, fill =Peach!15,thick]
\tikzstyle{SLU_model} = [rectangle,rounded corners=1mm,minimum width = 1.6cm, minimum height = 0.6cm, text centered, draw = SeaGreen, fill = SeaGreen!15,thick]
\tikzstyle{input} = [rectangle,rounded corners=0mm,minimum width = 4.3cm, minimum height = 0.65cm, text centered, draw = Teal, fill = CornflowerBlue!15,thick]
\tikzstyle{output} = [rectangle,rounded corners=0mm,minimum width = 4.3cm, minimum height = 0.65cm, text centered, draw = Thistle, fill = Thistle!15,thick]

\tikzstyle{word} = [rectangle,rounded corners=0.5mm,minimum width = 1.5cm, minimum height = 0.75cm, text centered, draw = SkyBlue, fill = CornflowerBlue!15,thick]
\tikzstyle{slot} = [rectangle,rounded corners=0.5mm,minimum width = 1.5cm, minimum height = 0.75cm, text centered, draw = Peru, fill = Peach!15,thick]

\tikzstyle{wrong} = [rectangle,rounded corners=0.25mm,minimum width = 4cm, minimum height = 1.15cm, draw =red, dashed]
\tikzstyle{wrong2} = [rectangle,rounded corners=0.25mm,minimum width = 4cm, minimum height = 1.15cm, draw =red, thick]

\begin{tikzpicture}[node distance = 0cm]

\tikzstyle{every node}=[scale=0.7]
\node(CENTER){};
\node(word1)[word,below of=CENTER,xshift=-2.5cm,yshift=1.7cm]{Return};
\node(word2)[word,right of=word1,xshift=2cm,yshift=0]{New};
\node(word3)[word,right of=word2,xshift=2cm,yshift=0]{York};
\node(word4)[word,right of=word3,xshift=2cm,yshift=0]{at};
\node(word5)[word,right of=word4,xshift=2cm,yshift=0]{9};
\node(word6)[word,right of=word5,xshift=2cm,yshift=0]{o'clock};

\node(wslot1)[slot,below of=word1,xshift=0cm,yshift=-1.2cm]{O};
\node(wslot2)[slot,right of=wslot1,xshift=2cm,yshift=0]{B-city};
\node(wslot3)[slot,right of=wslot2,xshift=2cm,yshift=0]{I-city};
\node(wslot4)[slot,right of=wslot3,xshift=2cm,yshift=0]{O};
\node(wslot5)[slot,right of=wslot4,xshift=2cm,yshift=0]{B-time};
\node(wslot6)[slot,right of=wslot5,xshift=2cm,yshift=0]{I-time};
\node()[below of=word4,xshift=-1cm,yshift=1cm]{\textbf{Non-autoregressive model}:};

\node(word1_2)[word,below of=wslot1,xshift=0cm,yshift=4.5cm]{Return};
\node(word2_2)[word,right of=word1_2,xshift=2cm,yshift=0]{New};
\node(word3_2)[word,right of=word2_2,xshift=2cm,yshift=0]{York};
\node(word4_2)[word,right of=word3_2,xshift=2cm,yshift=0]{at};
\node(word5_2)[word,right of=word4_2,xshift=2cm,yshift=0]{9};
\node(word6_2)[word,right of=word5_2,xshift=2cm,yshift=0]{o'clock};
\node()[below of=word4_2,xshift=-1cm,yshift=1cm]{\textbf{Autoregressive model}:};

\node(slot1)[slot,below of=word1_2,xshift=0cm,yshift=-1.2cm]{O};
\node(slot2)[slot,right of=slot1,xshift=2cm,yshift=0]{B-city};
\node(slot3)[slot,right of=slot2,xshift=2cm,yshift=0]{I-city};
\node(slot4)[slot,right of=slot3,xshift=2cm,yshift=0]{O};
\node(slot5)[slot,right of=slot4,xshift=2cm,yshift=0]{B-time};
\node(slot6)[slot,right of=slot5,xshift=2cm,yshift=0]{I-time};




\draw[-latex,thin] (word1.south) -- (wslot1.north); 
\draw[-latex,thin] (word2.south) -- (wslot2.north); 
\draw[-latex,thin] (word3.south) -- (wslot3.north);
\draw[-latex,thin] (word4.south) -- (wslot4.north); 
\draw[-latex,thin] (word5.south) -- (wslot5.north); 
\draw[-latex,thin] (word6.south) -- (wslot6.north); 

\draw[-latex,thin] (word1_2.south) -- (slot1.north); 
\draw[-latex,thin] (word2_2.south) -- (slot2.north); 
\draw[-latex,thin] (word3_2.south) -- (slot3.north);
\draw[-latex,thin] (word4_2.south) -- (slot4.north); 
\draw[-latex,thin] (word5_2.south) -- (slot5.north); 
\draw[-latex,thin] (word6_2.south) -- (slot6.north);

\draw[-latex,thin](word1_2.east)  -- (word2_2.west);
\draw[-latex,thin](word2_2.east)  -- (word3_2.west);
\draw[-latex,thin](word3_2.east)  -- (word4_2.west);
\draw[-latex,thin](word4_2.east)  -- (word5_2.west);
\draw[-latex,thin](word5_2.east)  -- (word6_2.west);

\end{tikzpicture}
\caption{An example of autoregressive model and non-autoregressive. Autoregressive model generates outputs word by word from left-to-right direction. Non-autoregressive model can produce outputs in parallel.}
\label{fig:example_auto}
\end{figure}
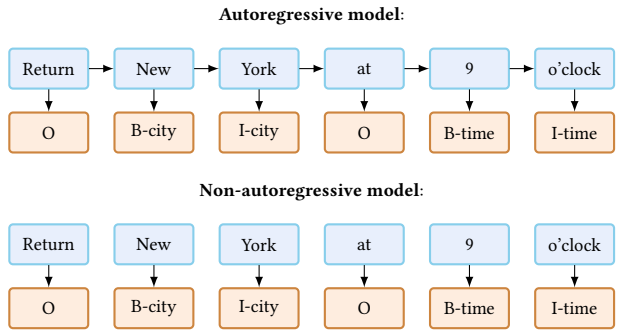

An intuitive idea for accelerating the predicting process is to utilize the self-attention (i.e., Transformer) \cite{Attention_is_all_you_need} instead of the RNN as the basic framework in a non-autoregressive way.
However, utilizing the Transformer-based model in multi-turn SLU tasks will face the following problems:

\textbf{Missing salient information:}
To deal with the multi-turn dialogue, many existing Transformer-based models \cite{zeng2020meddialog,zhou-etal-2021-generation} simply concatenate all historical utterances and the current utterance together. However, those approaches cause severe problems with increasing dialogue turns. First, simply concatenating sentences creates redundant computations. More seriously, concatenating all utterances together will form a very long sequence. The long sequence may cause the model difficulty to focus on the salient information due to the self-attention mechanism, resulting in a low SLU performance. Therefore, how to design an efficient module to process dialogue history without missing salient information is a major challenge.

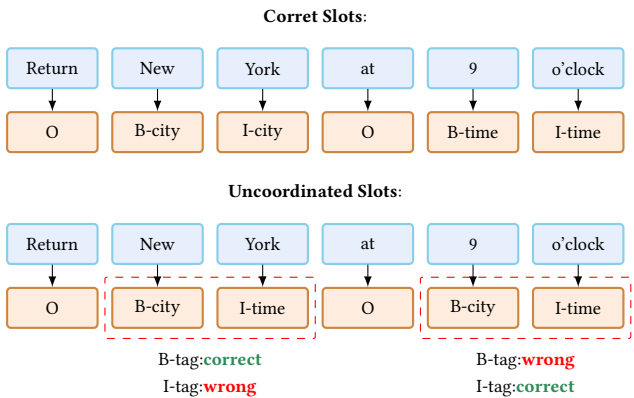
\begin{figure}[h]
\centering
\tikzstyle{background1} = [rectangle, rounded corners=3mm,minimum width = 14cm, minimum height = 5cm, text centered, draw = black, fill = gray!8]
\tikzstyle{memory_list1} = [rectangle,rounded corners=1mm,minimum width = 4cm, minimum height = 0.6cm, text centered, draw = Peru, fill =Peach!15,thick]
\tikzstyle{memory_list2} = [rectangle,rounded corners=1mm,minimum width = 4cm, minimum height = 1.5cm, text centered, draw = Peru, fill =Peach!15,thick]
\tikzstyle{SLU_model} = [rectangle,rounded corners=1mm,minimum width = 1.6cm, minimum height = 0.6cm, text centered, draw = SeaGreen, fill = SeaGreen!15,thick]
\tikzstyle{input} = [rectangle,rounded corners=0mm,minimum width = 4.3cm, minimum height = 0.65cm, text centered, draw = Teal, fill = CornflowerBlue!15,thick]
\tikzstyle{output} = [rectangle,rounded corners=0mm,minimum width = 4.3cm, minimum height = 0.65cm, text centered, draw = Thistle, fill = Thistle!15,thick]

\tikzstyle{word} = [rectangle,rounded corners=0.5mm,minimum width = 1.72cm, minimum height = 0.75cm, text centered, draw = SkyBlue, fill = CornflowerBlue!15,thick]
\tikzstyle{slot} = [rectangle,rounded corners=0.5mm,minimum width = 1.72cm, minimum height = 0.75cm, text centered, draw = Peru, fill = Peach!15,thick]

\tikzstyle{wrong} = [rectangle,rounded corners=0.25mm,minimum width = 4cm, minimum height = 1.15cm, draw =red, dashed]
\tikzstyle{wrong2} = [rectangle,rounded corners=0.25mm,minimum width = 4cm, minimum height = 1.15cm, draw =red, thick]

\begin{tikzpicture}[node distance = 0cm]

\tikzstyle{every node}=[scale=0.7]
\node(CENTER){};
\node(word1)[word,below of=CENTER,xshift=-2.5cm,yshift=1.7cm]{Return};
\node(word2)[word,right of=word1,xshift=2cm,yshift=0]{New};
\node(word3)[word,right of=word2,xshift=2cm,yshift=0]{York};
\node(word4)[word,right of=word3,xshift=2cm,yshift=0]{at};
\node(word5)[word,right of=word4,xshift=2cm,yshift=0]{9};
\node(word6)[word,right of=word5,xshift=2cm,yshift=0]{o'clock};

\node(wslot1)[slot,below of=word1,xshift=0cm,yshift=-1.2cm]{O};
\node(wslot2)[slot,right of=wslot1,xshift=2cm,yshift=0]{B-city};
\node(wslot3)[slot,right of=wslot2,xshift=2cm,yshift=0]{I-time};
\node(wslot4)[slot,right of=wslot3,xshift=2cm,yshift=0]{O};
\node(wslot5)[slot,right of=wslot4,xshift=2cm,yshift=0]{B-city};
\node(wslot6)[slot,right of=wslot5,xshift=2cm,yshift=0]{I-time};
\node()[below of=word4,xshift=-1cm,yshift=1cm]{\textbf{Uncoordinated Slots}:};

\node(word1_2)[word,below of=wslot1,xshift=0cm,yshift=4.5cm]{Return};
\node(word2_2)[word,right of=word1_2,xshift=2cm,yshift=0]{New};
\node(word3_2)[word,right of=word2_2,xshift=2cm,yshift=0]{York};
\node(word4_2)[word,right of=word3_2,xshift=2cm,yshift=0]{at};
\node(word5_2)[word,right of=word4_2,xshift=2cm,yshift=0]{9};
\node(word6_2)[word,right of=word5_2,xshift=2cm,yshift=0]{o'clock};
\node()[below of=word4_2,xshift=-1cm,yshift=1cm]{\textbf{Corret Slots}:};

\node(slot1)[slot,below of=word1_2,xshift=0cm,yshift=-1.2cm]{O};
\node(slot2)[slot,right of=slot1,xshift=2cm,yshift=0]{B-city};
\node(slot3)[slot,right of=slot2,xshift=2cm,yshift=0]{I-city};
\node(slot4)[slot,right of=slot3,xshift=2cm,yshift=0]{O};
\node(slot5)[slot,right of=slot4,xshift=2cm,yshift=0]{B-time};
\node(slot6)[slot,right of=slot5,xshift=2cm,yshift=0]{I-time};


\node(wrong_I)[wrong,below of=wslot2,xshift=1cm]{};
\node(PI1)[below of=wrong_I,yshift=-1cm]{
{B-tag:\color{SeaGreen} \textbf{correct}}
};
\node(PI2)[below of=wrong_I,yshift=-1.5cm]{
{I-tag:\color{red} \textbf{wrong}}
};

\node(wrong_B)[wrong,below of=wslot5,xshift=1cm]{};
\node(PB2)[below of=wrong_B,yshift=-1cm]{
{B-tag:\color{red} \textbf{wrong}}
};
\node(PB2)[below of=wrong_B,yshift=-1.5cm]{
{I-tag:\color{SeaGreen} \textbf{correct}}
};

\draw[-latex,thin] (word1.south) -- (wslot1.north); 
\draw[-latex,thin] (word2.south) -- (wslot2.north); 
\draw[-latex,thin] (word3.south) -- (wslot3.north);
\draw[-latex,thin] (word4.south) -- (wslot4.north); 
\draw[-latex,thin] (word5.south) -- (wslot5.north); 
\draw[-latex,thin] (word6.south) -- (wslot6.north); 

\draw[-latex,thin] (word1_2.south) -- (slot1.north); 
\draw[-latex,thin] (word2_2.south) -- (slot2.north); 
\draw[-latex,thin] (word3_2.south) -- (slot3.north);
\draw[-latex,thin] (word4_2.south) -- (slot4.north); 
\draw[-latex,thin] (word5_2.south) -- (slot5.north); 
\draw[-latex,thin] (word6_2.south) -- (slot6.north); 
\end{tikzpicture}
\caption{An example of uncoordinated slot problem.}
\label{fig:example-uc}
\end{figure}

\textbf{Uncoordinated slot problem:} Directly utilizing the Transformer framework will reduce the prediction performance of the SF task in each single turn \cite{wu2020slotrefine}. The reason is that SF is a sequence labeling task whose results utilize the “Inside–Outside–Beginning (IOB)” tagging format. Therefore, SF heavily depends on the strongness of the sequential dependency information among each slot chunk (proved by \citet{ren2020studyNAT}). However, as shown in Fig. \ref{fig:example_auto} the slot labels are predicted independently and simultaneously in non-autoregressive methods, which reduces the strongness of the sequential dependency and causes the uncoordinated slot problem. Fig. \ref{fig:example-uc} shows an example of the uncoordinated slot problem, where \texttt{B-city} should be followed by \texttt{I-city}, but \texttt{I-time} is predicted by mistake. 

Although \citet{wu2020slotrefine} tries to handle the uncoordinated slot problem via a two-pass mechanism called SlotRefine, this method still has the following weakness:
First, SlotRefine supposes all the uncoordinated slot problems are caused by wrong `\emph{I-tags}' and only utilize `\emph{B-tags}' to correct the `\emph{I-tags}' in its second pass period. 
However, as shown in Fig. \ref{fig:example-uc}, when predicting, an incorrect `\emph{B-tag}' follows by a correct `\emph{I-tag}' also happens. In this case, SlotRefine cannot utilize `\emph{I-tags}' to correct the `\emph{B-tags}'.
Second, the two-pass mechanism needs to characterize text and predict two times in both the training and testing process, which inevitably leads to lower efficiency.
Therefore, how to maintain the non-autoregressive efficiency when inferring while avoiding the uncoordinated problem in SLU tasks is a significant challenge.

To solve the above challenges, in this paper, we propose a novel non-autoregressive model for multi-turn SLU called Salient History Attention with Layer-Refined Transformer (\modelname{}). Our \modelname{} contains a Salient History Attention (SHA) module, a Transformer encoder with Layer-Refined Mechanism (LRM), and a Slot Label Generation (SLG) auxiliary task. Specifically, the SHA is proposed to capture salient historical information for the first challenge, where a history-attention approach is designed based on the self-attention mechanism. The query (\textbf{Q}), key (\textbf{K}), and value (\textbf{V}) of our history-attention is generated by the current utterance, historical utterances, and historical results, respectively. Since the predicted results of SLU in dialogue history contain essential and specific semantics (proved by\citet{clzICME2021}), our SHA module could obtain the hidden states of dialogue history with salient historical information. Subsequently, we utilize the hidden states of dialogue history to guide the prediction of the current-turn utterance. Utilizing calculated hidden states but not all utterances for the current utterance prediction avoids redundant computation. Thus our SHA module is quite efficient.

To further solve the uncoordinated problem for the second challenge, we propose LRM and SLG, respectively. Specifically, LRM modifies the hidden states between Transformer layers according to the intermediate predicted ID and SF results to help the final prediction. In practice, LRM predicts the preliminary results of ID and SF by the hidden states of the former Transformer layer, merging the result embeddings with the hidden states and inputting them into the next Transformer layer. We have further proved in experiments that LRM only needs to be employed once and does not need to be used between every Transformer layer, so the cost is much smaller than running the entire model twice. SLG is an auxiliary task that predicts the next SF label according to the utterance sequence and the generated SF labels, which is a sequence-to-sequence (seq2seq) task like machine translation. We jointly train the SLG and the original sequence labeling-based SLU tasks by multi-task learning and share t encoder. By SLG, the shared encoder learns more sequential dependency information from the decoder through the cross attention mechanism in the Transformer and further improves sequence labeling-based SLU tasks. Notable, SLG is only employed in the training process and consumes no extra inference time. Thus our model is still a fast non-autoregressive model.


We conduct extensive experiments to evaluate the effect of our complete \modelname{}. Specifically, we first evaluate our model on the multi-turn SLU dataset KVRET. The experiments results indicate that our model significantly improves multi-turn SLU performance (16.1\% on SF F1 score, 17.5\% on Overall) while substantially speeding up (nearly 15 times) the inference process over the state-of-the-art (SOTA) baselines. We further evaluate our model on two single-turn SLU datasets (without the SHA module). The results show that our model still obtains the best performance and the fastest speed compared with the current SOTA single-turn SLU models, which also manifests that SHA is a portable module and our whole model works robustly. We also make the ablation study and verify the effeteness of each module of our model. Finally, we combine our model with the pre-trained language models, and 
the results imply that the pre-trained language models could further boost the performance of our \modelname{}.

The main contributions of this paper are presented as follows:
\begin{enumerate}
\item We propose the SHA, which takes advantage of the dialogue history efficiently without losing the salient information and evidently enhances the SLU performance for the current utterance.
\item We design the SLG as auxiliary multitasking of SLU, which increases the sequential dependency information of the model to modify the uncoordinated problem while consuming no extra inference time.
\item We propose the LRM, which improves the overall performance of SLU tasks by the interaction of predicted results between Transformer layers with a negligible time cost.
\item Experimental results on three public datasets show our model is superior to both existing SOTA autoregressive and non-autoregressive models in terms of speed and performance, indicating that our model has great potential for real-world application.
\end{enumerate}

\section{Related Work}
\label{sec: Related Work}
In this section, we introduce the related work from two aspects.

\subsection{Single-turn SLU}
In SLU, intent detection is usually seen as a semantic classification problem to predict the intent label, and slot filling is mainly regarded as a sequence labeling task.
With the developments in deep neural networks, many methods \cite{yao2014lstm,mesnil2014rnn,peng2015rnn,kurata2016lstm} have been proposed to solve these two tasks. Traditionally, pipeline approaches are utilized to manage the two mentioned tasks separately. These kinds of methods typically suffer from error propagation due to their independent models and prove less effective than the joint models.
Motivated by this problem, some work deeply researches the relationship between these two tasks and finds that ID and SF are closely related. \citet{Slot-gated2018} propose a slot-gated model that learns the relationship between intent and slot by the gate mechanism. Inspired by \cite{Slot-gated2018}, some bi-directional networks are proposed \cite{bi-dictional2019,liu2019cm}, which dig into the correlation between ID and SF deeper and model the relationship between them more explicitly. In the above work, the interrelated connections between ID and SF are established. Besides, \citet{JointCapsule2019} utilize a hierarchical capsule neural network structure encapsulating the hierarchical relationship among utterance, slot, and intent. \citet{qin2019stack} propose a Stack-Propagation framework that uses result information of ID to guide the SF task but ignores the impact of SF results on the ID task. Although Stack-Propagation reaches the SOTA performance on two SLU datasets, it surfers a long inference latency caused by the heavy and complex framework. \citet{qin2020co} then concern about the impact of SF results in their model but still suffering the long inference latency. Inspired by \cite{yang2019}, \citet{clzICME2021} propose a portable framework RPFSLU, which contains a two-round predicting period. RPFSLU utilizes the semantic information of the first round prediction results to guide the second round prediction by a represent learning process. However, although RPFSLU is portable, it is not a lightweight model due to its two-round prediction process.

The above methods mostly use the autoregressive models (e.g., LSTM and GRU \cite{gru}) that suffer inevitable long inference latency. Inspired by the good performance of Transformer in other tasks 
\cite{liu2020regularized,li2021ode,li2021learning}, \citet{wu2020slotrefine} then propose a non-autoregressive joint model named SlotRefine for SLU, which speeds up the predicting process and encounters the uncoordinated-slot problem. SlotRefine tries to handle this problem by a two-pass refine mechanism and get some effect. However, it still suffers uncoordinated slot problems caused by the wrong `I-tag' and is not efficient enough during inference due to its two-pass mechanism that needs to run the whole model twice. 

Meanwhile, some work \cite{Multi-domainJoint2016,Multi-TaskNetworks2019} tries to enhance the performance via multi-task learning. 
These multi-task learning methods link the two tasks implicitly via applying a joint loss function.
\citet{memory2019} propose an impressive multi-task learning approach for multi-turn SLU by consolidating context memory with a dialogue logistic inference task called DLI. DLI needs no extra labeled data and only needs to be carried out during training, which inspired us to design SLG.

Compared with previous work, our \modelname{} model obtains the sequential dependency information via auxiliary multitasking and utilizes the interaction of the prediction results between ID and SF via the Transformer's middle states. 
Thus, our model has high predicting accuracy and fast inference speed, which indicates that our model has great potential for industrial application.

\subsection{Multi-turn SLU}
In multi-turn dialogue SLU, the model is provided with a group of extra context-sensitive dialog statements \cite{TOIS2021Survey}. To deal with the contextual information, many models incorporating more context from dialogue history \cite{End-to-end2016} or semantic context from the frame \cite{bapna2017towards} are proposed and outperformed the methods without context information. Using dialogue context is shown to gain improvement for an end-to-end dialogue \cite{bordes2016e2e}. Besides, \citet{sequentialSlu2017} proposed an RNN based model SDEN to improve dialogue context modeling and \cite{memory2019} refined SDEN with SDEN+, which proposed a dialogue logistic inference task for contextual SLU without needing extra labeled data. Then context-aware GCN \cite{qin2021knowing} is proposed to automatically aggregate the contextual information, which frees the SLU model from the manually designed heuristic aggregation function. All these models only concern context utterance but ignore the historical results, which contain abundant contextual information.

\citet{clzICME2021} then propose RPFSLU, which attain contextual information from both historical utterance and historical predicted results. Meanwhile, RPFSLU is a portable framework that can easily let nearly all RNN-based single-turn SLU models also be available in multi-turn SLU tasks. However, with the widespread of Transformer \cite{Attention_is_all_you_need}, Transformer-based models outperform RNN-based models on both accuracy and latency. Due to its basic architecture, RPFSLU is not compatible with Transformer-based SLU models. Compared with RPFSLU, our model is specially designed for Transformer-based SLU models and performs better on both speed and accuracy.

\section{Method}
\label{sec: Method}
In this section, we first introduce the problem formulation of multi-turn SLU and the overview of our \modelname{}. Then we describe our SHA module in detail. 
Subsequently, we describe the detail of our basic model and the Layer-Refined Mechanism. Finally, we introduce our Slot Label Generation task. 

\subsection{Problem Formulation}
In this section, we introduce the problem formulation for the multi-turn SLU task.

In multi-turn SLU tasks, dialogue system needs to make multiple turn of dialogue continuously. Therefore, in the $T-th$ turn, the input of SLU tasks are composed by current utterance $\textbf{X}=\{x_{cls},x_1,...,x_n\}$ and dialogue history $\textbf{DH} = \{ \textbf{X}^1,..., \textbf{X}^{T-1}\}$, where $n$ denotes the sequence length, $x_{cls}$ denotes the `\emph{CLS}' token, $\textbf{X}^{t}$ $(1 \leq t \leq T-1)$ denotes the utterance of the $t-th$ turn dialogue, respectively.

Given $\textbf{X}$ and $\textbf{DH}$ as input, our tasks are composed of Intent Detection (ID) and Slot Filling (SF). 
Specifically, ID is a semantic classification task to predict the intent label for the whole utterance, while SF is a sequence labeling task to give each token in the sequence a slot label. In our model, the intent label and all slot labels are predicted simultaneously.

\subsection{Overview}
In this section, we describe the overview of \modelname{} and introduce the relationship between each module in our whole model. 

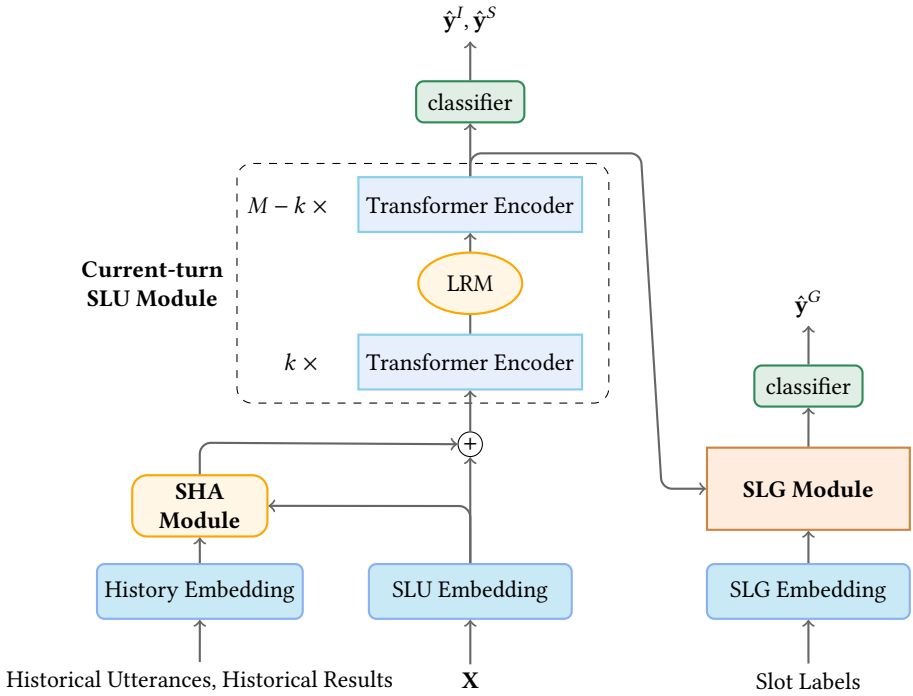
\begin{figure}[h]
\centering
\tikzstyle{background1} = [rectangle, rounded corners=3mm,minimum width = 7cm, minimum height = 3cm, text centered, draw = black, fill = gray!8]
\tikzstyle{background2} = [rectangle, rounded corners=2mm,minimum width = 5.5cm, minimum height = 3.5cm, text centered, draw = black, dashed]

\tikzstyle{Embedding} = [rectangle,rounded corners=1mm,minimum width = 3cm, minimum height = 0.8cm, text centered, draw = CornflowerBlue!75, fill = SkyBlue!45,thick]
\tikzstyle{Encoder} = [rectangle,rounded corners=0mm,minimum width = 3cm, minimum height = 0.8cm, text centered, draw = SkyBlue, fill = CornflowerBlue!15,thick]
\tikzstyle{Decoder} = [rectangle,minimum width = 3cm, minimum height = 1.2cm, text centered, draw = Peru, fill = Peach!15,thick]
\tikzstyle{SHA} = [rectangle,rounded corners=2mm, minimum height = 0.8cm,minimum width = 2cm, text centered, draw = Orange,fill = Orange!10,thick]
\tikzstyle{LRM} = [ellipse,inner sep=2mm, text centered, draw = Orange,fill = Orange!10,thick]
\tikzstyle{classifier} = [rectangle,rounded corners=1mm,minimum width = 1.6cm, minimum height = 0.6cm, text centered, draw = SeaGreen, fill = SeaGreen!15,thick]
\tikzstyle{cat} = [circle,minimum size =0.3cm,inner sep=0pt, text centered, draw = black]
\tikzstyle{add} = [circle,minimum size =0.3cm,inner sep=0pt, text centered, draw = black]
\tikzstyle{blank2} = [rectangle,rounded corners=0mm, minimum width = 0.2cm, minimum height = 0.2cm, text centered, thick]

\begin{tikzpicture}[node distance = 0cm,scale=9/6]
\tikzstyle{every node}=[scale=0.9]

\node(Embedding_e)[Embedding]{SLU Embedding};
\node(input_e)[below of=Embedding_e,yshift=-1.3cm]{\textbf{X}};
\node(Embedding_h)[Embedding,below of=Embedding_e,xshift=-4cm]{History Embedding};
\node(input_h)[below of=Embedding_h,yshift=-1.3cm]{Historical Utterances, Historical Results};
\node(SHA)[SHA,below of=Embedding_h,yshift=1.25cm,align=center]{\textbf{SHA}\\\textbf{Module}};
\node(add)[add,below of=SHA,xshift=4cm,yshift=0.9cm,scale=1.2]{+};

\node(Encoder)[Encoder,below of=add,yshift=1.2cm]{Transformer Encoder};
\node()[right of=Encoder,xshift=-2.5cm]{$k\ \times$};
\node(LRM)[LRM,below of=Encoder,xshift=0cm,yshift=1.15cm]{LRM};

\node(Encoder2)[Encoder,below of=LRM,yshift=1.15cm]{Transformer Encoder};
\node()[right of=Encoder2,xshift=-2.7cm]{$M-k\ \times$};

\node(SLU)[background2, below of=LRM,xshift=-0.7cm]{};
\node()[right of=SLU,xshift=-4cm,align = center] {\textbf{Current-turn}\\\textbf{SLU Module}};

\node(liner_e)[classifier,below of=Encoder2,yshift=1.5cm]{classifier};
\node(y_slu)[below of=liner_e,yshift=1.2cm]{$\hat{\textbf{y}}^{I},\hat{\textbf{y}}^{S}$};

\node(Embedding_d)[Embedding,below of=Embedding_e,xshift=5cm]{SLG Embedding};
\node(input_d)[below of=Embedding_d,yshift=-1.3cm]{Slot Labels};
\node(Decoder)[Decoder,below of=Embedding_d,yshift=1.5cm]{\textbf{SLG Module}};

\node(liner_d)[classifier,below of=Decoder,yshift=1.5cm]{classifier};
\node(y_slg)[below of=liner_d,yshift=1.2cm]{$\hat{\textbf{y}}^{G}$};

\draw[->,thick,Black!60](input_e.north) -- (Embedding_e.south);
\draw[->,thick,Black!60](input_h.north) -- (Embedding_h.south);
\draw[->,thick,Black!60](Embedding_h.north) -- (SHA.south);
\draw[->,thick,Black!60](Embedding_e.north) 
-- ([yshift=0.4cm]Embedding_e.north) arc(0:90:0.1cm)
-- (SHA.east);
\node(SHA_north_p)[blank2,below of = SHA, yshift=0.885cm]{};
\draw[->,thick,Black!60](SHA.north) -- (SHA_north_p.south) arc(180:90:0.1cm) -- (add.west);

\draw[->,thick,Black!60](Embedding_e.north) -- (add.south);
\draw[->,thick,Black!60](add.north) -- (Encoder.south);
\draw[-,thick,Black!60](Encoder.north) -- (LRM.south);
\draw[->,thick,Black!60](LRM.north) -- (Encoder2.south);

\draw[->,thick,Black!60](Encoder2.north) -- (liner_e.south);
\draw[->,thick,Black!60](liner_e.north) -- (y_slu.south);

\draw[->,thick,Black!60](Encoder2.north) 
--([yshift=0.1cm]Encoder2.north) arc(180:90:0.1cm)
--([xshift=1.42cm,yshift=0.2cm]Encoder2.north) arc (90:0:0.1cm)
--([xshift=-0.4cm,yshift=0.1cm]Decoder.west)arc (180:270:0.1cm)
--(Decoder.west);

\draw[->,thick,Black!60](input_d.north) -- (Embedding_d.south);
\draw[->,thick,Black!60](Embedding_d.north) -- (Decoder.south);
\draw[->,thick,Black!60](Decoder.north) -- (liner_d.south);
\draw[->,thick,Black!60](liner_d.north) -- (y_slg.south);

\end{tikzpicture}
\caption{General framework.}
\label{fig:model}
\end{figure}

The general framework of our model is shown in Fig. \ref{fig:model}, where $M$ is the number of Transformer Encoder layers, and $k$ indicates we employ LRM in the interval of the $k$-th and the $(k+1)$-th Encoder layer. Generally, our \modelname{} consists of three parts, i.e., a Salient History Attention (SHA) module, a current-turn SLU module, and a Slot Label Generation (SLG) auxiliary task.

Specifically, we first employ the SHA module to deal with the dialogue history. The input of SHA is composed of three parts: the current utterance, historical utterances, and previously predicted results.
In SHA, we first employ self-attention to address the current utterance. Then, we utilize the history-utterance-attention and history-result-attention, respectively, to obtain the contextual representation of the current utterance with historical information. Note that the history-utterance-attention and history-result-attention merge the semantic information contained in the historical utterances and results (previously predicted SLU results) based on the influence of each historical utterance on the current utterance, respectively. Thus, the obtained representation of the current utterance is with salient historical information, and we use it as the input of our current-turn SLU module.
We will introduce the detail of SHA in section \ref{sec:SHA}.

Then, we employ the current-turn SLU module to predict the SLU results for the current utterance. Our current-turn SLU module processes the output of SHA with a $M$ layers Transformer Encoder to calculate ID and SF results. Particularly, to solve the uncoordinated slot problem, we utilize Layer-Refined Mechanism (LRM) between two specific ($k$ and $k+1$) Transformer layers. In LRM, we predict the preliminary results of ID and SF by the hidden states of the $k$-th Transformer layer, merging the result embeddings with the hidden states and inputting them into the $(k+1)$-th Transformer layer. 
We will describe the detail of the Transformer Encoder and LRM in section \ref{sec:LRM}.

Finally, we introduce an auxiliary task, i.e., Slot Label Generation (SLG) 
to further solve the uncoordinated slot problem. The SLG extends the current-turn SLU module with a Transformer decoder (called SLG module) and forms a seq2seq task in an autoregressive way. Specifically, the SLG module aims to generate each of the current-turn slot labels based on previously generated slot labels to obtain sequential dependency information for our current-turn SLU module by the joint loss function. We will describe the detail of SLG in section \ref{sec:SLG}.

Notably, both the SHA and SLU modules work in a non-autoregressive way. Furthermore, the SLG task is only carried out during the training process, which costs no extra time for inference. Therefore, our complete model is an effective non-autoregressive SLU model.

\subsection{Salient History Attention}
\label{sec:SHA}
In this section, we proposed the Salient History Attention (SHA) module to address the salient information of dialogue history.

\textbf{History Embedding}:
Given the current utterance $\textbf{X} = \{x_1,...,x_n\}$ and dialogue history $\textbf{DH} = \{\textbf{X}_1,...,\textbf{X}_{T-1}\}$, we first embed current utterance $\textbf{X}$ into $\textbf{e} = \{ \textbf{e}_{cls},\textbf{e}_1,...,\textbf{e}_n \in \mathbb{R}^{d_e} \}$ ($d_e$ is the embedding size) and each historical utterance $\textbf{X}^t$ $(1 \leq t \leq T-1)$ into 
$\textbf{e}^t = \{ \textbf{e}^t_{cls},\textbf{e}^t_1,...,\textbf{e}^t_{n_t} \in \mathbb{R}^{d_e} \}$ ($n_t$ is the length of $\textbf{X}^t$ ). Then we concatenate each $\textbf{e}^t$ together to obtain the historical utterance embedding sequence $\textbf{E}^{u} = \{ \textbf{e}^1;...;\textbf{e}^{T-1} \}$.

Moreover, when solving the $T-th$ turn dialogue, we have predicted and recorded the SLU results of $\textbf{X}^t$ $(1 \leq t \leq T-1)$, i.e.,
ID label $resI^{t}$ and SF labels $\textbf{resS}^{t} = \{resS^{t}_1,...,resS^{t}_{n_t} \}$, where $\textbf{resS}^{t}_j$ denotes the slot label of the $j$-th token $x_j$ in $\textbf{X}^t$. Similarly to what we do for utterances, we embed each $\textbf{resI}^t$ and $\textbf{resS}^t$ into $\textbf{eI}^t \in \mathbb{R}^{d_e}$ and $\textbf{eS}^t = \{ \textbf{eS}^t_1,...,\textbf{eS}^t_{n_t}  \in \mathbb{R}^{d_e} \}$ to obtain the historical result embedding sequence $\textbf{E}^{r} = \{ \textbf{eI}^1;\textbf{eS}^1;...;\textbf{eI}^{T-1};\textbf{eS}^{T-1} \}$.   


\textbf{History-Attention}:
As we mentioned, dialogue history needs to be significantly concerned in multi-turn SLU tasks. Actually, the historical utterance with higher semantic similarity to the current dialogue should have a more significant influence \cite{TOISLi2021Dialogue}.
Moreover, \citet{clzICME2021} proved that the predicted results in dialogue history also contain important semantics and are even more specific than utterances in SLU tasks. Therefore, both historical utterances and historical results should be considered to obtain salient information from dialogue history.

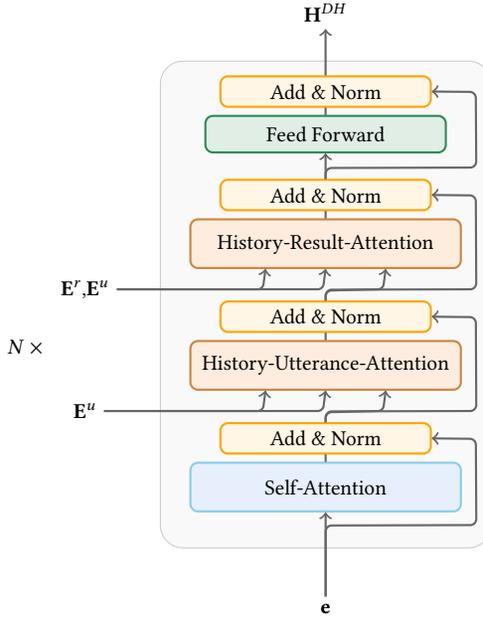
\begin{figure}[h]
\centering
\tikzstyle{background1} = [rectangle, rounded corners=3mm,minimum width = 5.5cm, minimum height = 8cm, text centered, draw = black, fill = gray!8]
\tikzstyle{background2} = [rectangle, rounded corners=3mm,minimum width = 7cm, minimum height = 8cm, text centered, draw = black, fill = gray!8]

\tikzstyle{Embedding} = [rectangle,rounded corners=1mm,minimum width = 2cm, minimum height = 0.8cm, text centered, draw = CornflowerBlue!75, fill = SkyBlue!45,thick]
\tikzstyle{Encoder} = [rectangle,rounded corners=1mm,minimum width = 4.5cm, minimum height = 0.8cm, text centered, draw = SkyBlue, fill = CornflowerBlue!15,thick]
\tikzstyle{Decoder} = [rectangle,rounded corners=1mm,minimum width = 4.5cm, minimum height = 0.8cm, text centered, draw = Peru, fill = Peach!15,thick]
\tikzstyle{norm} = [rectangle,rounded corners=1mm,minimum width = 3.5cm, minimum height = 0.5cm, text centered, draw = Orange,fill = Orange!10,thick]
\tikzstyle{LRM} = [ellipse,inner sep=2mm, text centered, draw = Orange,fill = Orange!10,thick]
\tikzstyle{FFN} = [rectangle,rounded corners=1mm,minimum width = 4cm, minimum height = 0.6cm, text centered, draw = SeaGreen, fill = SeaGreen!15,thick]
\tikzstyle{add} = [circle,minimum size =0.3cm,inner sep=0pt, text centered, draw = black]
\tikzstyle{blank} = [rectangle,rounded corners=0mm, minimum width = 1cm, minimum height = 0.8cm, text centered, thick]
\tikzstyle{blank2} = [rectangle,rounded corners=0mm, minimum width = 0.2cm, minimum height = 0.2cm, text centered, thick]

\begin{tikzpicture}[node distance = 0cm,scale=0.8]
\tikzstyle{every node}=[scale=0.8]
\node(middle)[blank2]{};
\node(background1)[background1, below of=middle, xshift=-3cm, fill=gray!5,draw=gray!40]{};
\node()[right of=background1,xshift=-5cm,yshift=-0.7cm]{$N\ \times$};
\node()[right of=background1,xshift=5cm,yshift=-0.7cm]{};

\node(Self_Attention)[Encoder,below of=background1,yshift=-3cm]{Self-Attention};
\node(input_e)[below of=Self_Attention,yshift=-2cm]{$\textbf{e}$};
\node(norm1)[norm,below of=Self_Attention,yshift=0.8cm]{Add \& Norm};
\node(his_Attention1)[Decoder,below of=norm1,yshift=1.2cm]{History-Utterance-Attention};
\node(Q1)[blank,below of=his_Attention1,xshift=1cm]{};
\node(K1)[blank,below of=his_Attention1,xshift=0cm]{};
\node(K1_input)[below of=K1,xshift=-4cm,yshift=-0.75cm]{$\textbf{E}^u$};
\node(V1)[blank,below of=his_Attention1,xshift=-1cm]{};
\node(norm2)[norm,below of=his_Attention1,yshift=0.8cm]{Add \& Norm};

\node(his_Attention2)[Decoder,below of=norm2,yshift=1.2cm]{History-Result-Attention};
\node(Q2)[blank,below of=his_Attention2,xshift=1cm]{};
\node(K2)[blank,below of=his_Attention2,xshift=0cm]{};
\node(K2_input)[below of=K2,xshift=-4cm,yshift=-0.75cm]{$\textbf{E}^r$,$\textbf{E}^u$};
\node(V2)[blank,below of=his_Attention2,xshift=-1cm]{};
\node(norm3)[norm,below of=his_Attention2,yshift=0.8cm]{Add \& Norm};
\node(ffn)[FFN,below of=norm3,yshift=1cm]{Feed Forward};
\node(norm4)[norm,below of=ffn,yshift=0.7cm]{Add \& Norm};
\node(A_o)[below of=norm4,yshift=1.3cm]{$\textbf{H}^{DH}$};

\draw[->,thick,Black!60](input_e.north) -- (Self_Attention.south);
\draw[-,thick,Black!60](Self_Attention.north) -- (norm1.south);

\node(norm1_c0)[blank2,below of=Self_Attention,xshift=0cm,yshift=-0.6cm]{};
\node(norm1_c1)[blank2,below of=Self_Attention,xshift=2.275cm,yshift=-0.625cm]{};
\node(norm1_c2)[blank2,below of=norm1,xshift=2.5cm,yshift=0.025cm]{};
\draw[->, thick, Black!60](input_e.north)  -- (norm1_c0.south)arc(180:90:0.1cm) -- (norm1_c1.east)arc(270:360:0.1cm) -- (norm1_c2.south)arc(0:90:0.1cm)
--(norm1.east);

\node(Q1_c0)[blank2,below of=norm1,xshift=0cm,yshift=0.2cm]{};
\node(Q1_c)[blank2,below of=Q1,xshift=0cm,yshift=-0.75cm]{};
\draw[->, thick, Black!60](norm1.north)  -- (Q1_c0.north)arc(180:90:0.1cm) -- (Q1_c.west)arc(270:360:0.1cm) -- (Q1.south);
\node(K1_c)[blank2,below of=K1,xshift=0cm,yshift=-0.75cm]{};
\draw[->, thick, Black!60](K1_input.east) 
--(K1_c.west) arc(270:360:0.1cm) -- (K1.south);
\node(V1_c)[blank2,below of=V1,xshift=0cm,yshift=-0.75cm]{};
\draw[->, thick, Black!60](K1_input.east) 
--(V1_c.west) arc(270:360:0.1cm) -- (V1.south);
\draw[-,thick,Black!60](his_Attention1.north) -- (norm2.south);

\node(norm2_c0)[blank2,below of=his_Attention1,xshift=0cm,yshift=-0.725cm]{};
\node(norm2_c1)[blank2,below of=his_Attention1,xshift=2.275cm,yshift=-0.75cm]{};
\node(norm2_c2)[blank2,below of=norm2,xshift=2.5cm,yshift=0.025cm]{};
\draw[->, thick, Black!60](norm1.north)  -- (norm2_c0.south)arc(180:90:0.1cm) -- (norm2_c1.east)arc(270:360:0.1cm) -- (norm2_c2.south)arc(0:90:0.1cm)
--(norm2.east);

\node(Q2_c0)[blank2,below of=norm2,xshift=0cm,yshift=0.2cm]{};
\node(Q2_c)[blank2,below of=Q2,xshift=0cm,yshift=-0.75cm]{};
\draw[->, thick, Black!60](norm2.north)  -- (Q2_c0.north)arc(180:90:0.1cm) -- (Q2_c.west)arc(270:360:0.1cm) -- (Q2.south);
\node(K2_c)[blank2,below of=K2,xshift=0cm,yshift=-0.75cm]{};
\draw[->, thick, Black!60](K2_input.east) 
--(K2_c.west) arc(270:360:0.1cm) -- (K2.south);
\node(V2_c)[blank2,below of=V2,xshift=0cm,yshift=-0.75cm]{};
\draw[->, thick, Black!60](K2_input.east) 
--(V2_c.west) arc(270:360:0.1cm) -- (V2.south);
\draw[-,thick,Black!60](his_Attention2.north) -- (norm3.south);

\node(norm3_c0)[blank2,below of=his_Attention2,xshift=0cm,yshift=-0.725cm]{};
\node(norm3_c1)[blank2,below of=his_Attention2,xshift=2.275cm,yshift=-0.75cm]{};
\node(norm3_c2)[blank2,below of=norm3,xshift=2.5cm,yshift=0.025cm]{};
\draw[->, thick, Black!60](norm2.north)  -- (norm3_c0.south)arc(180:90:0.1cm) -- (norm3_c1.east)arc(270:360:0.1cm) -- (norm3_c2.south)arc(0:90:0.1cm)
--(norm3.east);

\draw[->,thick,Black!60](norm3.north) -- (ffn.south);
\draw[-,thick,Black!60](ffn.north) -- (norm4.south);

\node(norm4_c0)[blank2,below of=ffn,xshift=0cm,yshift=-0.525cm]{};
\node(norm4_c1)[blank2,below of=ffn,xshift=2.275cm,yshift=-0.55cm]{};
\node(norm4_c2)[blank2,below of=norm4,xshift=2.5cm,yshift=0.025cm]{};
\draw[->, thick, Black!60](norm3.north)  -- (norm4_c0.south)arc(180:90:0.1cm) -- (norm4_c1.east)arc(270:360:0.1cm) -- (norm4_c2.south)arc(0:90:0.1cm)
--(norm4.east);

\draw[->,thick,Black!60](norm4.north) -- (A_o.south);

\end{tikzpicture}
\caption{SHA architecture.}
\label{fig:model_his}
\end{figure}

Motivated by the above requirements, we design a non-autoregressive architecture called SHA based on the Transformer Encoder.
Specifically, we insert two history-attention sub-layers, i.e., the history-utterance-attention sub-layer and the history-result-attention sub-layer between the self-attention sub-layer and the feed-forward network of the Transformer encoder. The history-utterance-attention and history-result-attention are utilized to obtain semantic information contained in the historical utterances and results, respectively. Besides, we employ residual connections \cite{he2016deep} around each of the sub-layers, followed by layer normalization \cite{norm}. 
Similar to Transformer, SHA is an N-layer architecture, and each layer has the same structure.
A brief view of SHA architecture is shown in Fig. \ref{fig:model_his}.

Before introducing our SHA module, we first briefly introduce the attention mechanism used in the Transformer. The attention function takes a query matrix $\textbf{Q} \in \mathbb{R}^{d_q }$, a key matrix $\textbf{K} \in \mathbb{R}^{d_k }$ and a value matrix $\textbf{V} \in \mathbb{R}^{d_v}$ as input, and is defined as
\begin{align}
    \operatorname{Attention}(x,&y,z) = \operatorname{softmax}\left(\frac{\textbf{Q}  \textbf{K}^{T}}{\sqrt{d_{k}}}\right) \textbf{V}\\
    &\begin{aligned}
        &\textbf{Q} = \mathrm{FC}(\textbf{x})\\
        &\textbf{K} = \mathrm{FC}(\textbf{y})\\
        &\textbf{V} = \mathrm{FC}(\textbf{z})
    \end{aligned}
\end{align}
where $\mathrm{FC}$ is the linear project operation.

In our SHA, we first employ the self-attention to encode the current utterance sequence by
\begin{align}
    \tilde{\textbf{H}}^c = &\operatorname{Attention}(\textbf{e},\textbf{e},\textbf{e})\\
    \textbf{H}^c = &\operatorname{Norm}(\textbf{e}+\tilde{\textbf{H}}^c)
\end{align}
where $\textbf{H}^c= \{\textbf{h}^{c}_{cls},\textbf{h}^{c}_1,...,\textbf{h}^{c}_n \in \mathbb{R}^{d_{model}}  \}$ denotes the hidden states of the current utterance,
$d_{model}$ is the input and output dimension of the SHA layer, and $\operatorname{Norm}()$ denotes the layer normalization. In this paper, we set $d_{model}=d_e$. Particularly, in the $j$-th ($2 \leq j \leq N$) SHA layer, we replace $\textbf{e}$  with the output of the $(j-1)$-th SHA layer as the input of self-attention.

To merge the semantic information of historical utterances into the current utterance based on the semantic similarity, we design a history-utterance-attention over current utterance and historical utterances by
\begin{align}
 \tilde{\textbf{H}}^u = &\operatorname{Attention}(\textbf{H}^c,\textbf{E}^u,\textbf{E}^u)\\
 \textbf{H}^u = &\operatorname{Norm}(\textbf{H}^c+\tilde{\textbf{H}}^u)
\end{align}

To further incorporate the salient information contained in historical results, we inject historical results information into the current utterance according to the semantic similarity between their corresponding historical utterance and the current utterance, and 
design the history-result-attention by
\begin{align}
 \tilde{\textbf{H}}^r = &\operatorname{Attention}(\textbf{H}^u,\textbf{E}^u,\textbf{E}^r)\\
 \textbf{H}^r = &\operatorname{Norm}(\textbf{H}^u+\tilde{\textbf{H}}^r)
\end{align}

We then employ a feed-forward network (FFN) to prevent the vanishing or exploding of gradients by
$\textbf{H}^{DH}= \{\textbf{h}^{DH}_{cls},\textbf{h}^{DH}_1,...,\textbf{h}^{DH}_n \in \mathbb{R}^{d_{model}}  \}$ by 
\begin{align}
    \tilde{\textbf{H}}^{DH} &= \operatorname{FFN}(\textbf{E}^r)\\
    \textbf{H}^{DH} &= \operatorname{Norm}(\textbf{E}^r+\tilde{\textbf{H}}^{DH})
\end{align}
where FFN is consists of two linear transformations with a ReLU activation function in between.



The SHA module finally returns $\hat{\textbf{H}}^{DH} = \{\hat{\textbf{H}}_{cls},\hat{\textbf{H}}^{DH}_1,...,\hat{\textbf{H}}^{DH}_n \in \mathbb{R}^{d_{model}}  \}$ from the output of the $N$-th SHA layer, which denotes the contextual representation of the current utterance. 

Subsequently, We merge $\hat{\textbf{H}}^{DH}$ with the embedding sequence of the current utterance $\textbf{e}$, and obtain the input of our current-utterance SLU module $\hat{\textbf{e}} = \{  \hat{\textbf{e}}_{cls}, \hat{\textbf{e}}_1,..., \hat{\textbf{e}}_n\in \mathbb{R}^{d_{model}}\}$ by 
\begin{align}
    \hat{\textbf{e}}  = \hat{\textbf{H}}^{DH} + \textbf{e}
\label{eq: cat}
\end{align}

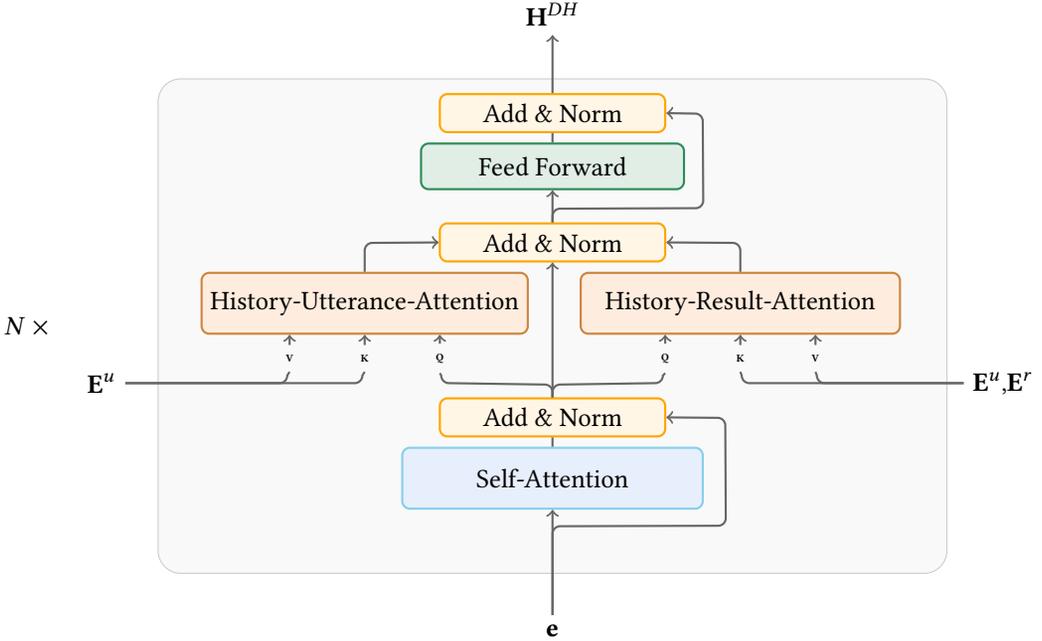
\begin{figure}[h]
\centering
\tikzstyle{background1} = [rectangle, rounded corners=3mm,minimum width = 10.5cm, minimum height = 6.5cm, text centered, draw = black, fill = gray!8]
\tikzstyle{background2} = [rectangle, rounded corners=3mm,minimum width = 7cm, minimum height = 8cm, text centered, draw = black, fill = gray!8]

\tikzstyle{Embedding} = [rectangle,rounded corners=1mm,minimum width = 2cm, minimum height = 0.8cm, text centered, draw = CornflowerBlue!75, fill = SkyBlue!45,thick]
\tikzstyle{Encoder} = [rectangle,rounded corners=1mm,minimum width = 4cm, minimum height = 0.8cm, text centered, draw = SkyBlue, fill = CornflowerBlue!15,thick]
\tikzstyle{Decoder} = [rectangle,rounded corners=1mm,minimum width = 4.25cm, minimum height = 0.8cm, text centered, draw = Peru, fill = Peach!15,thick]
\tikzstyle{norm} = [rectangle,rounded corners=1mm,minimum width = 3cm, minimum height = 0.5cm, text centered, draw = Orange,fill = Orange!10,thick]
\tikzstyle{LRM} = [ellipse,inner sep=2mm, text centered, draw = Orange,fill = Orange!10,thick]
\tikzstyle{FFN} = [rectangle,rounded corners=1mm,minimum width = 3.5cm, minimum height = 0.6cm, text centered, draw = SeaGreen, fill = SeaGreen!15,thick]
\tikzstyle{add} = [circle,minimum size =0.3cm,inner sep=0pt, text centered, draw = black]
\tikzstyle{blank} = [rectangle,rounded corners=0mm, minimum width = 1cm, minimum height = 0.8cm, text centered, thick]
\tikzstyle{blank2} = [rectangle,rounded corners=0mm, minimum width = 0.2cm, minimum height = 0.2cm, text centered, thick]

\begin{tikzpicture}[node distance = 0cm,scale=1]
\tikzstyle{every node}=[scale=1]
\node(background1)[background1, below of=middle, xshift=0cm, fill=gray!5,draw=gray!40]{};
\node()[right of=background1,xshift=-7cm,yshift=-0.0cm]{$N\ \times$};
\node()[right of=background1,xshift=7cm,yshift=-0.0cm]{};

\node(Self_Attention)[Encoder,below of=background1,yshift=-2cm]{Self-Attention};
\node(input_e)[below of=Self_Attention,yshift=-2cm]{$\textbf{e}$};
\node(norm1)[norm,below of=Self_Attention,yshift=0.8cm]{Add \& Norm};

\node(his_Attention1)[Decoder,below of=norm1, xshift=-2.5cm, yshift=1.5cm] {History-Utterance-Attention};
\node(Q1)[blank,below of=his_Attention1,xshift=1cm]{};
\node(K1)[blank,below of=his_Attention1,xshift=0cm]{};
\node(K1_input)[below of=K1,xshift=-3.5cm,yshift=-1.05cm]{$\textbf{E}^u$};
\node(V1)[blank,below of=his_Attention1,xshift=-1cm]{};

\node(his_Attention2)[Decoder,below of=norm1, xshift=2.5cm,yshift=1.5cm] {History-Result-Attention};
\node(Q2)[blank,below of=his_Attention2,xshift=-1cm]{};
\node(K2)[blank,below of=his_Attention2,xshift=0cm]{};
\node(K2_input)[below of=K2, xshift=3.5cm,yshift=-1.05cm]{$\textbf{E}^u$,$\textbf{E}^r$};
\node(V2)[blank,below of=his_Attention2,xshift=1cm]{};

\node(norm23)[norm,below of=his_Attention1, xshift=2.5cm,yshift=0.8cm]{Add \& Norm};

\node(ffn)[FFN,below of=norm23,yshift=1cm]{Feed Forward};
\node(norm4)[norm,below of=ffn,yshift=0.7cm]{Add \& Norm};
\node(A_o)[below of=norm4,yshift=1.3cm]{$\textbf{H}^{DH}$};

\draw[->,thick,Black!60](input_e.north) -- (Self_Attention.south);
\draw[-,thick,Black!60](Self_Attention.north) -- (norm1.south);

\node(norm1_c0)[blank2,below of=Self_Attention,xshift=0cm,yshift=-0.6cm]{};
\node(norm1_c1)[blank2,below of=Self_Attention,xshift=2.075cm,yshift=-0.625cm]{};
\node(norm1_c2)[blank2,below of=norm1,xshift=2.3cm,yshift=0.025cm]{};
\draw[->, thick, Black!60](input_e.north)  -- (norm1_c0.south)arc(180:90:0.1cm) -- (norm1_c1.east)arc(270:360:0.1cm) -- (norm1_c2.south)arc(0:90:0.1cm)
--(norm1.east);

\node(Q1_c0)[blank2,below of=norm1,xshift=0cm,yshift=0.2cm]{};
\node(Q1_c)[blank2,below of=Q1,xshift=0.25cm,yshift=-1.05cm]{};
\node(Q1_t)[blank2,below of=Q1, xshift=0cm,yshift=-0.725cm, font=\fontsize{4}{6}\selectfont]{\textbf{Q}};
\draw[-, thick, Black!60](norm1.north)  -- (Q1_c0.north)arc(0:90:0.1cm) -- (Q1_c.west)arc(270:180:0.1cm) -- (Q1_t.south);
\draw[->, thick, Black!60](Q1_t.north) -- (Q1.south);
\node(K1_c)[blank2,below of=K1,xshift=0cm,yshift=-1.05cm]{};
\node(K1_t)[blank2,below of=K1, xshift=0cm,yshift=-0.725cm, font=\fontsize{4}{6}\selectfont]{\textbf{K}};
\draw[-, thick, Black!60](K1_input.east) 
--(K1_c.west) arc(270:360:0.1cm) -- (K1_t.south);
\draw[->, thick, Black!60](K1_t.north) -- (K1.south);
\node(V1_c)[blank2,below of=V1,xshift=0cm,yshift=-1.05cm]{};
\node(V1_t)[blank2,below of=V1, xshift=0cm,yshift=-0.725cm, font=\fontsize{4}{6}\selectfont]{\textbf{V}};
\draw[-, thick, Black!60](K1_input.east) 
--(V1_c.west) arc(270:360:0.1cm) -- (V1_t.south);
\draw[->, thick, Black!60](V1_t.north) -- (V1.south);

\node(Q2_c0)[blank2,below of=norm1,xshift=0cm,yshift=0.2cm]{};
\node(Q2_c)[blank2,below of=Q2,xshift=0cm,yshift=-1.05cm]{};
\node(Q2_t)[blank2,below of=Q2, xshift=0cm,yshift=-0.725cm, font=\fontsize{4}{6}\selectfont]{\textbf{Q}};
\draw[-, thick, Black!60](norm1.north)  -- (Q2_c0.north)arc(180:90:0.1cm) -- (Q2_c.west)arc(270:360:0.1cm) -- (Q2_t.south);
\draw[->, thick, Black!60](Q2_t.north) -- (Q2.south);
\node(K2_c)[blank2,below of=K2,xshift=0.25cm,yshift=-1.05cm]{};
\node(K2_t)[blank2,below of=K2, xshift=0cm,yshift=-0.725cm, font=\fontsize{4}{6}\selectfont]{\textbf{K}};
\draw[-, thick, Black!60](K2_input.west) 
--(K2_c.west) arc(270:180:0.1cm) -- (K2_t.south);
\draw[->, thick, Black!60](K2_t.north) -- (K2.south);
\node(V2_c)[blank2,below of=V2,xshift=0.25cm,yshift=-1.05cm]{};
\node(V2_t)[blank2,below of=V2, xshift=0cm,yshift=-0.725cm, font=\fontsize{4}{6}\selectfont]{\textbf{V}};
\draw[-, thick, Black!60](K2_input.west)
--(V2_c.west) arc(270:180:0.1cm) -- (V2_t.south);
\draw[->, thick, Black!60](V2_t.north) -- (V2.south);

\node(norm23_cl)[blank2,below of=norm23,xshift=-2.5cm,yshift=0.025cm]{};
\node(norm23_cr)[blank2,below of=norm23,xshift=2.5cm,yshift=0.025cm]{};
\draw[->,thick,Black!60](his_Attention1.north) -- (norm23_cl.south)arc(180:90:0.1cm)  -- (norm23.west);
\draw[->,thick,Black!60](his_Attention2.north) -- (norm23_cr.south)arc(0:90:0.1cm)  -- (norm23.east);

\draw[->,thick,Black!60](norm1.north) -- (norm23.south);
\draw[->,thick,Black!60](norm23.north) -- (ffn.south);
\draw[-,thick,Black!60](ffn.north) -- (norm4.south);

\node(norm4_c0)[blank2,below of=ffn,xshift=0cm,yshift=-0.525cm]{};
\node(norm4_c1)[blank2,below of=ffn,xshift=1.775cm,yshift=-0.55cm]{};
\node(norm4_c2)[blank2,below of=norm4,xshift=2cm,yshift=0.025cm]{};
\draw[->, thick, Black!60](norm23.north)  -- (norm4_c0.south)arc(180:90:0.1cm) -- (norm4_c1.east)arc(270:360:0.1cm) -- (norm4_c2.south)arc(0:90:0.1cm)
--(norm4.east);

\draw[->,thick,Black!60](norm4.north) -- (A_o.south);

\end{tikzpicture}
\caption{SHA-Parallel architecture.}
\label{fig:model_his2}
\end{figure}
\textbf{SHA-Parallel architecture:}
To accelerate the predicting process further, we can also operate two history-attention functions in parallel and merge their output. This architecture is called SHA-Parallel (SHA-P), as shown in Fig. \ref{fig:model_his2}. 

In SHA-P, the \textbf{Q} matrix of both two History-attention sub-layers are generated from $\textbf{H}^c$.
We calculate the contextual representation of the current utterance from both historical utterances and results in SHA-P by 
\begin{align}
    \textbf{H}^u &= \operatorname{Attention}(\textbf{H}^c,\textbf{E}^u,\textbf{E}^u)\\
    \textbf{H}^r &= \operatorname{Attention}(\textbf{H}^c,\textbf{E}^u,\textbf{E}^r)
\end{align}
and merge $\textbf{H}^u$ and $\textbf{H}^r$ by
\begin{align}
    \textbf{H}^{ur} &= \operatorname{Norm}(\textbf{H}^c+\textbf{H}^u+\textbf{H}^r)
\end{align}
where $\textbf{H}^{ur}$ is the merged contextual representation.

The contextual representation of the current utterance $\textbf{H}^{DH}$ in SHA-P is calculated by
\begin{align}
    \tilde{\textbf{H}}^{DH} &= \operatorname{FFN}(\textbf{H}^{ur})\\
    \textbf{H}^{DH} &= \operatorname{Norm}(\textbf{H}^{ur}+\tilde{\textbf{H}}^{DH})
\end{align}

Generally, SHA-P infers faster than SHA, while SHA has a better performance. We will further analyze the advantage and weaknesses of these two architectures in section \ref{sec: Experiments}. 

By our SHA module, $\hat{\textbf{e}}$ contains current utterance information together with the salient information of dialogue history. Particularly, SHA is a portable module. When solving single-turn SLU tasks, we can directly input $\textbf{e}$ into our current-utterance SLU module.

\subsection{Transformer Encoder with Layer-Refined Mechanism}
\label{sec:LRM}
In this section, we will introduce the detail of our current-turn SLU module.

\textbf{Basic model:}
Following previous non-autoregressive models \cite{wu2020slotrefine,cheng2021effective}, we employ an M-layer Transformer encoder to predict SLU results for current utterance, which is a non-autoregressive structure. To better capture the sequential information, we utilize the relative position representations \cite{shaw2018position_relative_attention} instead of the absolute position embedding used in the original Transformer.

Specifically, given $\hat{\textbf{e}} = \{ \hat{\textbf{e}}_{cls},\hat{\textbf{e}}_1,...,\hat{\textbf{e}}_n \}$ as input, we first employ a position relative self-attention to address the dependencies of each tokens by
\begin{align}
    &\begin{aligned}
        \beta_{pq} &= \frac{\hat{\textbf{e}}_{p} W^{Q}\left(\hat{\textbf{e}}_{q} W^{K}+a_{pq}^{K}\right)^{T}}{\sqrt{d_{k}}}\\
        \alpha_{pq}&=\frac{\exp \beta_{pq}}{\sum_{j=1}^{n} \exp \beta_{pj}}\\
        \textbf{Z}_{p} &= \sum_{q=1}^{n} {{\alpha}}_{pq}\left(\hat{\textbf{e}}_{q} \textbf{W}^{V}+{\textbf{a}}_{pq}^{V} \right)
    \end{aligned}
\label{eq: self-attention-basic}
\end{align}
where $\textbf{Z} = \{ \textbf{Z}_{cls},\textbf{Z}_{1}...,\textbf{Z}_{n} \in \mathbb{R}^{ d_{model} }\}$ is the self-attention output,
$\textbf{W}^Q \in \mathbb{R}^{d_{model} \times d_q}$, $\textbf{W}^K \in \mathbb{R}^{d_{model} \times d_k}$, and $\textbf{W}^V \in \mathbb{R}^{d_{model} \times d_v}$ are fully connected matrices, $d_q=d_k=d_v$ is equal to $d_{model}$, $\textbf{a}^K_{pq}$ and $\textbf{a}^V_{pq}$
indicates the relative position representations of token $x_p$ and $x_q$. Following \citet{shaw2018position_relative_attention}, 
we obtain $\textbf{a}^K_{pq}$ and $\textbf{a}^V_{pq}$ by
\begin{align}
    \begin{aligned}
        \textbf{a}^K_{pq} &= \textbf{w}^K_{clip(p-q,l)}\\
        \textbf{a}^V_{pq} &= \textbf{w}^V_{clip(p-q,l)}\\
        clip(x,l) &= max(-l,min(x,l))
    \end{aligned}
\end{align}
where $l$ is the farthest relative position,
$\textbf{w}^K = \{ \textbf{w}^K_{-l},...,\textbf{w}^K_{l} \in \mathbb{R}^{d_{model}} \} $, $\textbf{w}^V = \{ \textbf{w}^V_{-l},...,\textbf{w}^V_{l} \in \mathbb{R}^{d_{model}} \} $ are learnable vectors, which are initialized randomly and modified with the training process.

Notably, the $j$-th Transformer Encoder layer, we replace $\hat{\textbf{e}}$ with the output of the $(j-1)$-th ($2 \leq j \leq M$) Transformer Encoder layer as the input of the self-attention to calculate $\textbf{Z}$.

Subsequently, we employ a feed forward network (FFN) to attain the output of the $j$-th Transformer encoder layer $\textbf{H}^j = \{ \textbf{h}^j_{cls}, \textbf{h}^j_1,...,\textbf{h}^j_{n} \in \mathbb{R}^{ d_{model} }\}$ by
\begin{align}
    \tilde{\textbf{H}}^j &= FFN(\textbf{Z})\\
    \textbf{H}^j &= \operatorname{Norm}(\textbf{Z}+\tilde{\textbf{H}}^j)
\label{eq: ffn-basic}
\end{align}

The Transformer Encoder finally outputs $\textbf{H} = \{ \textbf{h}_{cls}, \textbf{h}_1,...,\textbf{h}_{n} \in \mathbb{R}^{ d_{model} }\}$ from the $M$-th Transformer Encoder layer ($\textbf{H}$ is actually $\textbf{H}^M$). Then, we utilize $\textbf{H}$ to calculate the prediction of ID and SF by
\begin{equation}
\begin{aligned}
    \hat{\textbf{y}}^I &= \mathrm{softmax} ( \textbf{W}^I  \textbf{h}_{cls} + \textbf{b}^I )\\
    \hat{\textbf{y}}^S_j &= \mathrm{softmax} ( \textbf{W}^S  (\textbf{h}_j \oplus \textbf{h}_{cls}) + \textbf{b}^S )
\end{aligned}
\label{eq:getres}
\end{equation}
where $\hat{\textbf{y}}^I \in \mathbb{R}^{d_{i}} $ and $\hat{\textbf{y}}^S = \{ \hat{\textbf{y}}^S_1,..., \hat{\textbf{y}}^S_n \in \mathbb{R}^{d_{s}} \}$ denote the predicted ID and SF result distributions of the current utterance, $d_{i}$ and $d_s$ are the the categories of the intent label and slot labels,
$\textbf{W}^I \in \mathbb{R}^{ d_{i} \times d_{model} }$ and $\textbf{W}^S \in \mathbb{R}^{ d_{s} \times 2d_{model}}$ are fully connected matrices, 
$\textbf{b}^I \in \mathbb{R}^{d_{i}}$ and $\textbf{b}^S \in \mathbb{R}^{d_{s}}$ are bias vectors, and $\oplus$ denotes the concatenation operation.

The objective of our basic model can be formulated as:
\begin{equation}
    P ({y}^{I}, {y}^{S} \mid \textbf{X}, \textbf{DH} ) =P(\textbf{y}^{I} \mid \textbf{X}, \textbf{DH}) \cdot \prod_{j}^{n} P(\textbf{y}_{j}^{S} \mid \textbf{X}, \textbf{DH}, \hat{\textbf{y}}^{I})
\end{equation}

And the joint loss function of SLU is defined as:
\begin{equation}
\begin{aligned}
\mathcal{L}_{SLU} &=-\log P({y}^{I} \mid \textbf{X}, \textbf{DH}) - \sum^{n}_{j=1} \log P({y}_{j}^{S} \mid \textbf{X}, \textbf{DH})
\end{aligned}
\end{equation}
Note that ${y}^{I}$ and ${y}^{S}_j$ are the ground truth labels (scalar), while $\hat{\textbf{y}}^I$ and $\hat{\textbf{y}}^S_j$ are the predicted result distributions whose dimensions are equal to the categories of their tasks.

Besides, during inference, we obtain the label of ID and SF by
\begin{equation}
\begin{aligned}
    resI &= \mathrm{argmax} ( \textbf{W}^I  \textbf{h}_{cls} + \textbf{b}^I )\\
    resS_j &= \mathrm{argmax} ( \textbf{W}^S  (\textbf{h}_j \oplus \textbf{h}_{cls}) + \textbf{b}^S )
\end{aligned}
\label{eq:GetSluLabel}
\end{equation}
where $resI$ and $\textbf{resS} = \{ {resS}_1,..., {resS}_n \}$ donate the predicted ID label and SF labels of the current utterance. We will record them and further utilize them in our SHA module as part of the historical results in the following turn (i.e., $T+1$ turn) of dialogue.

\textbf{Layer-Refined Mechanism (LRM):}
Previous work \cite{qin2019stack, clzICME2021} has widely proved that ID and SF are closely related. Taking advantage of the correlation between these two tasks, especially utilizing one task's results in the other, can effectively enhance the overall performance.
However, since the non-autoregressive approach predicts the results of ID and SF simultaneously, we can not directly employ these results in a one-pass prediction process like Stack-Propagation \cite{qin2019stack} do.
Although we can utilize a two-pass mechanism \cite{wu2020slotrefine} to generate the first-pass results and guide the second-pass prediction via them, it is a trade-off between autoregression and non-autoregression, which costs much time.
\begin{figure}[h]
\centering
\tikzstyle{background1} = [rectangle, rounded corners=3mm,minimum width = 7cm, minimum height = 3cm, text centered, draw = black, fill = gray!8]
\tikzstyle{background2} = [rectangle, rounded corners=3mm,minimum width = 7cm, minimum height = 8cm, text centered, draw = black, fill = gray!8]

\tikzstyle{blank} = [rectangle, rounded corners=1mm,minimum width = 5cm, minimum height = 0.8cm, draw = orange,  dashed]

\tikzstyle{Encoder} = [rectangle,rounded corners=2.5mm,minimum width = 8cm, minimum height = 0.8cm, text centered, draw = SkyBlue, fill = CornflowerBlue!15,thick]
\tikzstyle{Encoder_cell} = [rectangle,rounded corners= 0.5mm,minimum width = 1cm, minimum height = 0.6cm, text centered, draw = SkyBlue, fill = CornflowerBlue!15, thick]
\tikzstyle{Encoder_cell_b} = [rectangle,rounded corners= 0.5mm,minimum width = 1cm, minimum height = 0.8cm, text centered, thick]

\tikzstyle{Embedding} = [rectangle,rounded corners=1mm,minimum width = 7.2cm, minimum height = 0.8cm, text centered, draw = Orange, fill = Orange!10,thick]
\tikzstyle{Embedding_cell} = [rectangle,rounded corners=1mm,minimum width = 0.8cm, minimum height = 0.6cm, text centered, draw = Orange, fill = Orange!10,thick]
\tikzstyle{Embedding_cell_b} = [rectangle,rounded corners=1mm,minimum width = 1cm, minimum height = 0.8cm, text centered,thick]

\tikzstyle{Decoder} = [rectangle,minimum width = 2cm, minimum height = 0.8cm, text centered, draw = Peru, fill = Peach!15,thick]
\tikzstyle{LRM} = [ellipse,inner sep=2mm, text centered, draw = Orange,fill = Orange!10,thick]

\tikzstyle{classifier} = [rectangle,rounded corners=1mm,minimum width = 7.2cm, minimum height = 0.8cm, text centered, draw = SeaGreen, fill = SeaGreen!15,thick]
\tikzstyle{classifier_cell} = [rectangle,rounded corners=1mm,minimum width = 0.8cm, minimum height = 0.6cm, text centered, draw = SeaGreen, fill = SeaGreen!15, thick]
\tikzstyle{classifier_cell_b} = [rectangle,rounded corners=1mm,minimum width = 1cm, minimum height = 0.8cm, text centered, thick]

\tikzstyle{add} = [circle,minimum size =0.3cm,inner sep=0pt, text centered, draw = black!40]
\tikzstyle{add_b} = [circle,minimum size =0.3cm,inner sep=0pt, text centered]
\tikzstyle{Function} = [circle,minimum size =0.6cm,inner sep=0pt, text centered, draw = Orange,fill = Orange!10,thick]
\tikzstyle{blank1} = [rectangle,minimum width = 1cm, minimum height = 0.6cm, text centered, draw = white,draw opacity=0, fill = white, fill opacity=0,thick]

\begin{tikzpicture}[node distance = 0cm]
\tikzstyle{every node}=[scale=0.8]

\node(Encoder)[Encoder]{The $k$-th Encoder Layer };
\node(e_cls)[Encoder_cell_b,below of = Encoder,xshift=-3cm,yshift=0cm]{};
\node(e_1)[Encoder_cell_b,right of = e_cls,xshift=2cm,yshift=0cm]{}; 
\node(e_2)[Encoder_cell_b,right of = e_1,xshift=2cm,yshift=0cm]{}; 
\node(e_3)[Encoder_cell_b,right of = e_2,xshift=2cm,yshift=0cm]{};

\node(h_cls)[Encoder_cell, below of = e_cls,xshift=0cm,yshift=1.2cm]{$\textbf{h}^k_{cls}$};
\node(h_1)[Encoder_cell, right of = h_cls,xshift=2cm,yshift=0cm]{$\textbf{h}^k_1$}; 
\node(h_2)[Encoder_cell_b, right of = h_1,xshift=2cm,yshift=0cm]{...}; 
\node(h_3)[Encoder_cell, right of = h_2,xshift=2cm,yshift=0cm]{$\textbf{h}^k_n$};

\node(liner)[classifier,below of = h_cls,xshift=3cm,yshift=1.5cm]{Classifier};
\node(c0)[classifier_cell_b,below of = liner, xshift=-3cm]{};
\node(c1)[classifier_cell_b,below of = c0, xshift=2cm]{};
\node(c2)[classifier_cell_b,below of = c1, xshift=2cm]{};
\node(c3)[classifier_cell_b,below of = c2, xshift=2cm]{};

\node(y0)[classifier_cell,below of = c0, yshift=1.2cm]{$\tilde{\textbf{y}}^{I}$};
\node(y1)[classifier_cell,below of = y0, xshift=2cm]{$\tilde{\textbf{y}}^S_1$};
\node(y2)[classifier_cell_b,below of = y1, xshift=2cm]{...};
\node(y3)[classifier_cell,below of = y2, xshift=2cm]{$\tilde{\textbf{y}}^S_n$};

\node(Embedding)[Embedding,below of = y0,xshift=3cm,yshift=1.5cm]{Result Embedding};
\node(eb0)[Embedding_cell_b,below of = Embedding, xshift=-3cm]{};
\node(eb1)[Embedding_cell_b,below of = eb0, xshift=2cm]{};
\node(eb2)[Embedding_cell_b,below of = eb1, xshift=2cm]{};
\node(eb3)[Embedding_cell_b,below of = eb2, xshift=2cm]{};

\node(re0)[Embedding_cell,below of = eb0, yshift=1.2cm] {$\textbf{e}^{I},\textbf{e}_0^{S}$};
\node(re1)[Embedding_cell,below of = re0, xshift=2cm]{$\textbf{e}^S_1$};
\node(re2)[Embedding_cell_b,below of = re1, xshift=2cm]{...};
\node(re3)[Embedding_cell,below of = re2, xshift=2cm]{$\textbf{e}^S_n$};

\node(re123)[blank,below of = re2]{};


\node(add_I)[add,below of=re0,xshift=0cm,yshift=1.5cm,scale=1.2]{+};
\node(add1)[add_b,right of=add_I,xshift=2cm,yshift=0cm,scale=1.2]{...};
\node(add2)[right of=add1,xshift=2cm,yshift=0cm]{...};
\node(add3)[add,right of=add2,xshift=2cm,yshift=0cm,scale=1.2]{+};

\node(h_cls2)[Encoder_cell, below of = add_I,xshift=0cm,yshift=1.2cm]{${\textbf{h}'}_{cls}$};
\node(h_12)[Encoder_cell, right of = h_cls2,xshift=2cm,yshift=0cm]{$\textbf{h}'_1$}; 
\node(h_22)[Encoder_cell_b, right of = h_12,xshift=2cm,yshift=0cm]{...}; 
\node(h_32)[Encoder_cell, right of = h_22,xshift=2cm,yshift=0cm]{$\textbf{h}'_n$};

\node(Encoder2)[Encoder,below of=h_cls2,xshift=3cm,yshift=1.2cm]{The $(k+1)$-th Encoder Layer};
\node(e_cls2)[Encoder_cell_b,below of = Encoder2, xshift=-3cm, yshift=0cm]{};
\node(e_12)[Encoder_cell_b,right of = e_cls2, xshift=2cm, yshift=0cm]{}; 
\node(e_22)[Encoder_cell_b,right of = e_12,xshift=2cm,yshift=0cm]{}; 
\node(e_32)[Encoder_cell_b,right of = e_22, xshift=2cm, yshift=0cm]{};

\node(p0)[below of = h_cls, xshift=-1cm]{};
\node(p0_add)[below of = add_I, xshift=-1cm]{};
\draw[->, thin,SkyBlue](h_cls.west) -- (p0.east) -- (p0_add.east) --(add_I.west);


\node(p3)[below of = h_3, xshift=1cm]{};
\node(p3_add)[below of = add3, xshift=1cm]{};
\draw[->, thin,SkyBlue](h_3.east) -- (p3.west) -- (p3_add.west) --(add3.east);

\draw[-latex, thick,SkyBlue](e_cls.north)  -- (h_cls.south);
\draw[-latex, thick,SkyBlue](e_1.north)  -- (h_1.south);
\draw[-latex, thick,SkyBlue](e_3.north)  -- (h_3.south);

\draw[->, thick,SkyBlue](h_cls.north)  -- (c0.south);
\draw[->, thick,SkyBlue](h_1.north)  -- (c1.south);
\draw[->, thick,SkyBlue](h_3.north)  -- (c3.south);

\draw[->, dotted, thick, SkyBlue](h_cls.north)  .. controls ([xshift=0.2cm]h_cls.north) and ([xshift=-0.3cm,yshift=-0.5cm]c1.south) .. ([xshift=-0.1cm,yshift=-0.1cm]c1.south);

\draw[->, dotted, thick, SkyBlue](h_cls.north)  .. controls ([xshift=0.2cm]h_cls.north) and ([xshift=-0.8cm,yshift=-0.5cm]c2.south) .. ([xshift=-0.1cm,yshift=-0.1cm]c2.south);

\draw[->, dotted, thick, SkyBlue](h_cls.north)  .. controls ([xshift=0.2cm]h_cls.north) and ([xshift=-1cm,yshift=-0.5cm]c3.south) .. ([xshift=-0.1cm,yshift=-0.1cm]c3.south);

\draw[->,thick,SeaGreen](c0.north)  -- (y0.south);
\draw[->,thick,SeaGreen](c1.north)  -- (y1.south);
\draw[->,thick,SeaGreen](c3.north)  -- (y3.south);

\draw[->,thick,SeaGreen](y0.north)  -- (eb0.south);
\draw[->,thick,SeaGreen](y1.north)  -- (eb1.south);
\draw[->,thick,SeaGreen](y3.north)  -- (eb3.south);

\draw[->,thick,orange](eb0.north)  -- (re0.south);
\draw[->,thick,orange](eb1.north)  -- (re1.south);
\draw[->,thick,orange](eb3.north)  -- (re3.south);

\draw[->,thick,orange](re0.north)  -- (add_I.south);
\draw[->,thick,orange](re1.north)  -- (add1.south);
\draw[->,thick,orange](re3.north)  -- (add3.south);

\draw[->,thin,orange](re123.west)  -- (re0.east);

\draw[-latex,thick,Black!40](add_I.north) -- (h_cls2.south);
\draw[-latex,thick,Black!40](add1.north) -- (h_12.south);
\draw[-latex,thick,Black!40](add3.north) -- (h_32.south);

\draw[-latex,thick,SkyBlue](h_cls2.north) -- (e_cls2.south);
\draw[-latex,thick,SkyBlue](h_12.north) -- (e_12.south);
\draw[-latex,thick,SkyBlue](h_32.north) -- (e_32.south);

\end{tikzpicture}
\caption{LRM architecture.}
\label{fig:LRM-model}
\end{figure}
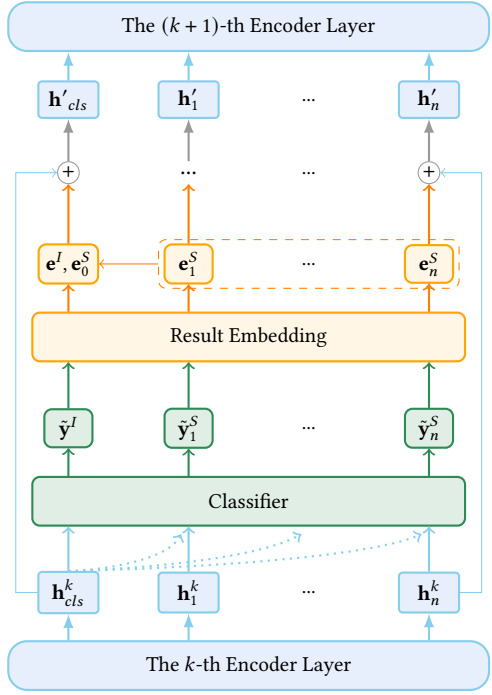

Considering that the Transformer contains a multi-layer architecture, and each encoder layer has the same structure. We design the LRM, which works between two Transformer layers and utilizes the middle states of the Transformer to guide the final prediction.

Specifically, we first predict the preliminary results of ID $\tilde{\textbf{y}}^I$ and SF $\tilde{\textbf{y}}^S=\{ \tilde{\textbf{y}}^S_1,...,\tilde{\textbf{y}}^S_n \}$ by Eq.\ref{eq:getres} according to hidden states $\textbf{H}^k=\{\textbf{h}^k_{cls},\textbf{h}^k_1,...,\textbf{h}^k_n \in \mathbb{R}^{d_{model}}\}$ from the $k$-th Transformer layer. 
Then, we convert $\tilde{\textbf{y}}^I$ into $\textbf{e}^{I} \in \mathbb{R}^{d_e}$, and $\tilde{\textbf{y}}^S=\{ \tilde{\textbf{y}}^S_1,...,\tilde{\textbf{y}}^S_n \}$ into $\textbf{e}^{S}=\{ \textbf{e}^S_1,...,\textbf{e}^S_n \in \mathbb{R}^{d_e}  \}$ by 
\begin{equation}
\begin{aligned}
    \textbf{e}^{I} &= \textbf{E}^{I}  \tilde{\textbf{y}}^I\\
    \textbf{e}^S_j &= \textbf{E}^{S}  \tilde{\textbf{y}}^S_j
\end{aligned}
\end{equation}
where $\textbf{E}^I \in \mathbb{R}^{d_e \times d_i}$ and $\textbf{E}^S \in \mathbb{R}^{d_e \times d_S}$ are the embedding metrics.  

Since the SF returns a sequence of results, we further calculate the weighted  average of their correponding embedding vectors via the attention mechanism by
\begin{equation}
    \textbf{e}^S_0=\sum_{j=1}^{n} \alpha_{j}  \textbf{e}^S_j
\end{equation}
where $\alpha_j$ is the weight of $\textbf{e}^S_j$ obtained by
\begin{equation}
    \alpha_{j}=\frac{
        \operatorname{exp}(\textbf{V} \cdot \textbf{e}^S_{j})}
        {\sum_{p=1}^{n} \operatorname{exp}( \textbf{V} \cdot \textbf{e}^S_{p})}
\end{equation}
$\textbf{e}^S_0 \in \mathbb{R}^{d_e}$ is the utterance-level SF result embedding vector, and $\textbf{V} \in \mathbb{R}^{d_e}$ is a trainable vector.

By the above calculation, we obtain result embedding vectors $\textbf{e}^{I}$, $\textbf{e}^{S}$ and $\textbf{e}^{S}_0$, which contain semantic information from the preliminary results of ID and SF. Subsequently, we merge these result embeddings with the former output and obtain a new hidden states sequence  $\textbf{H}'=\{\textbf{h}'_{cls},\textbf{h}'_1,...,\textbf{h}'_n \in \mathbb{R}^{d_{model}}\}$ by
\begin{align}
    \textbf{h}_{cls}'&= \textbf{h}^k_{cls} + \textbf{e}^{I} + \textbf{e}^{S}_0\\
    \textbf{h}_j'&= \textbf{h}^k_j+\textbf{e}^{S}_j
\end{align}
We use $\textbf{H}'$ as input of the $(k+1)$-th Transformer layer.

LRM can incorporate the bidirectional semantic information from one task to the other by propagating the combination of former output and preliminary results so that  ID and SF become more accurate.
The complete Markov chain process can be simplified as follow:
\begin{equation}
\begin{aligned}
    P(&{y}^{I}, {y}^{S}  \mid \textbf{X}, \textbf{DH}) =P(\tilde{\textbf{y}}^{I} \mid \textbf{X}, \textbf{DH}) \cdot  P(\tilde{\textbf{y}}^{S} \mid \textbf{X}, \textbf{DH}, \tilde{\textbf{y}}^{I})\\
    &\cdot  P(\textbf{y}^{I} \mid \textbf{X}, \textbf{DH}, \tilde{\textbf{y}}^{I}, \tilde{\textbf{y}}^{S} ) \cdot  P(\textbf{y}^{S} \mid \textbf{X}, \textbf{DH}, \tilde{\textbf{y}}^{I}, \tilde{\textbf{y}}^{S}, \hat{\textbf{y}}^{I} )
\end{aligned}
\end{equation}

LRM is a portable plugin for the Transformer, which can be utilized between two Transformer encoder layer intervals. Actually, using LRM once in the whole Transformer structure is enough because it already considers the interaction between ID and SF. On the contrary, overusing LRM  will make the Transformer layers lose their own feature and negatively affect the final performance because we use the SLU classifier in LRM. Also, LRM is not suitable for the following autoregressive SLG task because it simultaneously uses all SLU prediction labels.

Besides, the network structure of LRM is mainly composed of a fully connected layer as the SLU classifier and an embedding layer for result embedding. Therefore, our LRM is very lightweight and consumes very little inference time. 

\subsection{Slot Label Generation}
\label{sec:SLG}
In this section, we introduce the Slot Label Generation (SLG) task in detail.

As we aforementioned, the SF task heavily depends on the strongness of the sequential dependency information among each slot chunk. Although we have tried to obtain dependency information via self-attention with relative position representations \cite{shaw2018position_relative_attention}, it is not enough for the sequence labeling task like SF. 

To obtain more sequential dependency information for the model, we design the SLG as auxiliary multitasking, jointly training the original sequence labeling-based SLU tasks and sharing their encoders.

Specifically, given the token sequence
$\textbf{X} = \{ {x}_{cls}, {x}_1,...,{x}_{n} \}$ and previous slot labels $\{ {y}^G_{1},..., {y}^G_{j-1} \}$, SLG generates the next slot label $\textbf{y}^G_j$ by an autoregressive way. 
The objective of our SLG can be formulated as:
\begin{equation}
\begin{aligned}
    P( {y}^G \mid \textbf{X})= \prod_{j=1}^{n} P({y}_{j}^{G} \mid \textbf{X}, {y}^G_{1},...,{y}^G_{j-1}  )
\end{aligned}
\end{equation}
where ${y}^G = \{ {y}^G_1,..., {y}^G_n\}$ are the ground truth labels of SLG.

In practice, we share the encoder of SLU (our current-turn SLU module) to SLG task and extend the encoder with a Transformer decoder to construct the model architecture of SLG. 
The complete framework of SLG works as a seq2seq model like machine translation. Since the Transformer-based seq2seq model has been widely used, we do not describe it in more detail. Notably, following existing work, we utilize the teacher forcing strategy during the training process.

Moreover, to enhance the prediction consistency between the SLU and SLG tasks, we further design a consistency loss function based on the cross-entropy as
\begin{equation}
\begin{aligned}
\mathcal{L}_{SLG} 
&= (1-\alpha) \cdot -\log P( {y}^{G} \mid\textbf{X} ) +\alpha \cdot CE(  \hat{\textbf{y}}^{G} ,\hat{\textbf{y}}^{S} )\\
&= (\alpha-1) \cdot \log P( \textbf{y}^{G} \mid\textbf{X} ) +\alpha \cdot CE(  \hat{\textbf{y}}^{G} ,\hat{\textbf{y}}^{S} )
\label{eq_slgloss}
\end{aligned}
\end{equation}
where $CE$ is the cross-entropy function, $\alpha$ is a hyper-parameter, $\hat{\textbf{y}}^G = \{ \hat{\textbf{y}}^G_1,..., \hat{\textbf{y}}^G_n \in \mathbb{R}^{d_{s}} \}$ is the predicted distribution of SLG ,
and $\hat{\textbf{y}}^{S}$ is the predicted distribution of SF obtained by Eq.\ref{eq:getres}.  

By SLG, the shared encoder can learn more sequential dependency information from the decoder through the cross attention mechanism in Transformer and further improve sequence labeling-based SLU tasks. 

The loss function of overall multi-task learning is defined as:
\begin{equation}
    \mathcal{L}= \mathcal{L}_{SLU}+\lambda \mathcal{L}_{SLG}
\label{eq:jloss}
\end{equation}
where $\lambda$ is a hyper-parameter.

\section{Experiments}
\label{sec: Experiments}
In this section, we demonstrate the effectiveness of \modelname{}. 
We first introduce datasets, the necessary hyper-parameters, and the baselines used in our experiments.
Then, we compare the performance of our model with baselines and analyze the experiment results on multi-turn and single-turn SLU tasks, respectively.
Subsequently, we analyze the error caused by the uncoordinated problem and carry out an ablation study to verify the effeteness of SLG, LRM, and SHA.
Finally, we combine our model with the pre-trained model and analyze the effect.

\subsection{Experimental Settings and Baselines}

\paragraph{\textbf{Dataset:}}
To evaluate the efficiency of our proposed model, we conduct experiments on a multi-turn dataset: KVRET \cite{kvret}.
The dataset consists of 3,031 multi-turn dialogues, where 2,425 dialogues in the training set, 302 in the validation set and 304 in the test set. Each dialogue contains 5.25 turns on average.
More details of KVRET are shown in Table \ref{Table: dataset-multi}.
\begin{table}[h]
\centering
\begin{tabular}{lr}
\toprule
Dataset                 & KVRET\\
\midrule
Vocabulary Size         & 1601 \\
Avg. tokens per utterance    & 9.0     \\
Avg. turns per dialogue & 5.3  \\
Intent categories               & 3     \\
Slot categories               & 27     \\
Training set size             & 2425  \\
Validation set size           & 302   \\
Test set size                 & 304   \\
\bottomrule
\end{tabular}
\caption{Statistics for KVRET dataset.}
\label{Table: dataset-multi}
\end{table}

\paragraph{\textbf{Evaluation Metrics:}}
Following previous work, we utilize accuracy for ID and the F1 score for SF as the performance evaluation.
Besides, we utilize overall accuracy to indicate the proportion of utterance in the corpus whose slots and intent are both correctly predicted. Usually, a higher intent accuracy and F1 score also lead to higher overall accuracy. However, this does not always happen, e.g., when the prediction contains more mistakes, but most are from the same utterances.

\paragraph{\textbf{Setup:}}
We use AdamW \cite{adamw} to optimize the parameters in our model and adopted the suggested learning rate of 5e-5. The batch size is set to 32 according to the size of training data.

When tuning hyperparameters, we repeat the model 5 times and select the parameters with the best average performance on the validation set as the optimal. 

We first select the hyper-parameters used in our basic model. Specifically, we select $d_{model}$ in the range of $\{128,\ 256,\ 512,\ 768 \}$, and layers of SHA $N$ in $\{ 2,\ 3,\ 6 \}$ by the grid search. We finally choose $d_{model}$ as 768, and $N$ as 3. The size of the inner-layer in the feed-forward network $d_{ff}$ is set as four times of $d_{model}$. For other hyper-parameters , following \cite{Attention_is_all_you_need}, we set both encoder and decoder layers as 6, the number of attention heads as 8, and the dropout ratio as 0.3. We utilize LRM once between the second and the third Transformer layer.

Then, we choose the hyper-parameters $\alpha$ used in Eq.\ref{eq_slgloss} and $\lambda$ used in Eq.\ref{eq:jloss}. Specifically, we first fix $\lambda$ as 1 and select $\alpha$ in the range of (0,0.5] with the step 0.05. Subsequently, with the chosen $\alpha$, we select $\lambda$ in range of (0,1] with the step 0.25. We finally get the optimal when $\alpha$  is 0.35 and $\lambda$ is 0.75. We will introduce the influence of $\lambda$ and $\alpha$ in our ablation study.

\paragraph{\textbf{Baselines:}}
To evaluate the performance of our model, we select the following models as benchmark.

\textbf{Autoregressive Baselines}:
\begin{itemize}
\item MemNet \cite{End-to-end2016}: A model with attention-based memory retrieval.
\item SDEN \cite{sequentialSlu2017}: A model with sequential encoder based memory retrieval.
\item SDEN$^+$ \cite{memory2019}: An advanced version of SDEN using different RNN structure in decoder.
\item RPFSLU \cite{clzICME2021}: An portable framework that utilizes a GRU-based module to obtain historical information from historical utterances and results to help the current turn SLU task.
\item RPFSLU+BiModel \cite{birnn2018}: Utilizing BiModel as the basic model of RPFSLU, which achieves the best performance on ID and overall accuracy.
\item RPFSLU+Stack-Propagation \cite{qin2019stack}: Utilizing Stack-Propagation as the basic model of RPFSLU, which has the best performance on the SF task.
\end{itemize}

\textbf{Non-Autoregressive Baselines}:
\begin{itemize}
\item Basic model: The encoder of Transformer framework with relative position representations \cite{Attention_is_all_you_need}, which ignores historical information.
\item CAT-All: Dealing with dialogue history by concatenating all historical utterances with the current utterance together.
\item Basic model+CAT-All: Dealing with dialogue history by CAT-All, using the Basic model.
\end{itemize}

\textbf{Our models}:
\begin{itemize}
\item LRT: Our model without the SHA module, i.e., our current-turn SLU module with the SLG module.
\item LRT+CAT-All: Dealing with dialogue history by CAT-All, using the LRT model.
\item SHA-LRT: The complete model we proposed
\item SHA-P-LRT: SHA-LRT but utilizing the SHA-P to deal with dialogue history.
\end{itemize}

For Joint MemNet, SDEN, and SDEN$^+$, we adopt the reported results from \cite{clzICME2021}. For RPFSLU + BiModel and RPFSLU + Stack-Propagation, we re-implement the model. Note that, for all re-implemented models and our model, we repeat the experiment 5 times and report the average as the final results.

\subsection{Result and Analysis}
\label{section:Result}
In this section, we show the results of our experiments and do some analysis.

\paragraph{\textbf{SLU Performance:}}
\begin{table*}[ht]
\centering
\begin{tabular}{l|ccc}
\hline
\multirow{2}{*}{\textbf{Model}}  & \multicolumn{3}{c}{\textbf{KVRET}}     \\ 
\cline{2-4} & \multicolumn{1}{c}{Intent} & \multicolumn{1}{c}{Slot} & Overall \\
\hline
\multicolumn{4}{c}{\textbf{Autoregressive Models}} \\ 
\hline
MemNet \cite{End-to-end2016}       & 97.6  & 76.8  & 73.4  \\
SDEN \cite{sequentialSlu2017}          & 97.7  & 76.9  & 75.3  \\
SDEN$^+$ \cite{memory2019}          & 97.1  & 77.1  & 75.7  \\
RPFSLU \cite{clzICME2021}+Bi-model \cite{birnn2018} & 98.9 & 79.1 & 77.0 \\
RPFSLU \cite{clzICME2021}+Stack-Propagation \cite{qin2019stack} & 98.3 & 80.0 & 76.9\\
\hline
\multicolumn{4}{c}{\textbf{Non-autoregressive Models}} \\ 
\hline
Basic model  & 97.0      & 77.6     & 75.6 \\
Basic model + CAT-All & 94.4    & 78.0 	& 74.4 \\
LRT & 97.8 &78.3 &76.4 \\
LRT + CAT-All & 97.7 &78.0 &74.5 \\
\modelname{} & 99.0 & \ \textbf{96.1}$\uparrow$ & \ \textbf{94.4}$\uparrow$\\
SHA-P-LRT & \ \textbf{99.4}$\uparrow$ & 95.9 & 94.1\\
\hline
\end{tabular}
\caption{SLU performance on KVRET datasets.
The numbers with $\uparrow$ indicate that the improvement of our model over all baselines is statistically significant with $p < 0.05$ under t-test.}
\label{Table: res-multi}
\end{table*}

The experiment results on the KVRET dataset are shown in Table \ref{Table: res-multi}.
The results show that our model achieves the best performance in all three metrics and enhances baseline performance with an enormous margin. Specifically, compared with the prior SOTA model RPFSLU+Stack-Propagation, our model enhances the performance by 16.1\% on SF and 17.5\% on overall accuracy when utilizing SHA. Even with SHA-P, the enhancement is still 15.9\% on SF and 17.2\% on overall. 
This indicates the evident effectiveness of our \modelname{} on multi-turn SLU tasks. We attribute this enhancement to accurately catching the salient historical information by our SHA module.

As a comparison, simply concatenating all historical utterances perform unsatisfyingly. As shown in Table \ref{Table: res-multi}, the Basic model + CAT-All and LRT + CAT-All performs even worse than the Basic model and LRT, respectively. We attribute this phenomenon to the loss of salient information, which directly indicates the importance of capturing salient information when dealing with dialogue history.

Moreover, the performance of our Basic model and only utilizing our current-turn SLU module (LRT) are unsatisfactory due to lacking historical information, which directly indicates the importance of considering dialogue history in multi-turn SLU tasks. With SHA, the performance of these models is enhanced by a large margin (nearly 18\% on SF and Overall). We will further analyze the effect of SHA in more detail in section \ref{section: Ablation Study}.

\paragraph{\textbf{Speed Up:}}

\begin{table}[h]
\centering
\begin{tabular}{l|cc}
\hline
\multirow{2}{*}{Model} & \multicolumn{2}{c}{\textbf{KVRET}} \\ 
\cline{2-3} 
& \multicolumn{1}{c}{Latency } & Speedup \\ 
\hline
RPFSLU+Stack-Propagation        & 174.37ms   & 1.00$\times$  \\
RPFSLU+BiModel              & 125.35ms    & 1.39$\times$  \\ 
\hline
Basic model    & 6.99ms   & 24.95$\times$  \\
LRT    & 7.42ms   & 23.50$\times$  \\
\modelname{}   & 12.29ms   & 14.19$\times$  \\
SHA-P-LRT   & 11.96ms   & 14.58$\times$  \\
\hline
\end{tabular}
\caption{Latency of SLU models on multi-turn SLU tasks. ``Latency'' is the average inference time without minibatching. “Speedup” is compared against the existing SOTA model RPFSLU +Stack-Propagation.}
\label{Table: res_speed-multi}
\end{table}

The inference time of SLU models is shown in Table \ref{Table: res_speed-multi}. All the models in this experiment are conducted with a single TITAN Xp GPU. 

As shown in Table \ref{Table: res_speed-multi}, our proposed model achieves significant speedup against the autoregressive SOTA model RPFSLU+BiModel (11.31$\times$) and RPFSLU+Stack-Propagation (14.58$\times$). The main reason is that both the module for handling dialogue history and the module for processing single-turn SLU in our SHA-LRT work in a non-autoregressive way. Specifically, our SHA module deals with every turn simultaneously for dialogue history, while RPFSLU processes them turn by turn. Moreover, autoregressive models calculate each token in the current utterance serially while our model predicts SLU results for each token in parallel. Thus SHA works efficiently and much faster than RPFSLU.

As shown in Table \ref{Table: res_speed-multi}, the latency cost of SHA-LRT is only 4.87ms longer than that of the LRT, which indicates that our SHA module spends very little time. By the parallel structure, SHA-P works even 2.7\%  faster than SHA. Considering the significant enhancement on SLU performance brought by SHA, the trade-off on latency is acceptable. We also find that utilizing LRM consumes only 5.7\% extra inference time than without LRM, indicating that LRM is a lightweight approach with only a little time cost.

\subsection{Single-turn Performance}
\label{sec: Single-turn Performance}
As introduced in section \ref{sec: Method}, our SHA module is portable, which is also appropriate for single-turn SLU tasks. Therefore, we further conduct experiments to evaluate the performance of our model on single-turn SLU tasks. In this experiment, we use Adam \cite{adam} as the optimizer and adopted the suggested learning rate of 0.001. Besides, we set $d_{model}$ as 128. All other hyperparameters and the evaluation metrics are the same as we conduct in multi-turn SLU tasks. As the single-turn task contains no dialogue history, we only utilize the LRT part in our model, i.e., the current-turn SLU module with the SLG module. The embedding sequence of the current sequence is directly input to our current-turn SLU module.

\paragraph{\textbf{Dataset:}} We conduct experiments on two public datasets, i.e., ATIS (Airline Travel Information Systems \cite{atis}) and SNIPS (collected by Snips personal voice assistant \cite{snips}). Compared with ATIS, the SNIPS dataset is more complex due to its large vocabulary size, cross-domain intents, and more out-of-vocabulary words. The statistics of ATIS and SNIPS are shown in Table \ref{Table: dataset-single}.
\begin{table}[h]
\centering
\begin{tabular}{lrrr}
\toprule
Dataset                 & ATIS      & SNIPS     \\
\midrule
Vocabulary Size         & 722       & 11241     \\
Avg. tokens per utterance    & 11.28     & 9.05      \\
Intent categories             & 21        & 7         \\
Slot categories              & 120       & 72        \\
Training set size            & 4478      & 13084     \\
Validation set size          & 500       & 700       \\
Test set size                 & 893       & 700       \\
\bottomrule
\end{tabular}
\caption{Statistics for ATIS and SNIPS.}
\label{Table: dataset-single}
\end{table}

\paragraph{\textbf{Baselines:}}
To evaluate the performance of our model on single-turn SLU tasks, we compare our model with the existing baselines.

\textbf{Autoregressive Baselines:}
\begin{itemize}
\item Joint Seq \cite{Multi-domainJoint2016}: A GRU \cite{gru} based model with a multi-task modeling approach.
\item Attention-BiRNN \cite{attention-basedRnn2016}: A LSTM \cite{LSTM} based encoder-decoder model with an intent attention mechanism.
\item Slot-gated \cite{Slot-gated2018}: A LSTM based joint model together with a slot-gated mechanism as a special gate function.
\item Bi-model \cite{birnn2018}: An RNN based encoder-decoder model considering the intent and slot ﬁlling cross-impact to each other.
\item SF-ID \cite{bi-dictional2019}: A LSTM based joint model with cross-impact calculating between two tasks.
\item Stack-Propagation \cite{qin2019stack}: A LSTM based joint model with stack-propagation framework and token-level ID, which is the state-of-the-art of the autoregressive Single-Intent SLU model.
\item AG-IF \cite{qin2020agif}: A LSTM-based joint model with an adaptive interaction network, which is designed for Multi-Intent SLU tasks, and is also the state-of-the-art model on ATIS and SNIPS datasets.
\end{itemize}

\textbf{Non-autoregressive Baselines:}
\begin{itemize}
\item Basic model: The encoder of Transformer framework with relative position representations.  
\item GL-GIN \cite{qin2021glgin}: A LSTM-based joint model with a Global-Locally Graph Interaction Network, which is designed for Multi-Intent SLU task. 
\item SlotRefine \cite{wu2020slotrefine}: A Transformer based non-autoregressive model with a two-pass refine mechanism. 
\end{itemize}

For Joint Seq, Attention BiRNN, Slot-gated, BiModel, SF-ID, and Stack-Propagation, we adopt the reported results from \citet{qin2019stack}.
For SlotRefine, since the benchmark in their original paper is calculated nonstandardly according to their open-source code, we re-implemented the model (all hyper-parameters strictly identical as \citet{wu2020slotrefine}) and obtained the results. 
Since AG-IF and GL-GIN are specially designed for Multi-Intent SLU tasks, we re-implemented the model with the suggested hyper-parameters in their original paper.
Note that, for the re-implemented models, basic model, and our model, we repeat the experiment 5 times and report the average as the final results.

\begin{table*}[ht]
\centering
\resizebox{0.7\textwidth}{!}{
\begin{tabular}{l|ccc|ccc}
\hline
\multirow{2}{*}{\textbf{Model}}  & \multicolumn{3}{c|}{\textbf{ATIS}}    & \multicolumn{3}{c}{\textbf{SNIPS}} \\ 
\cline{2-7} & \multicolumn{1}{c}{Intent} & \multicolumn{1}{c}{Slot} & Overall & \multicolumn{1}{c}{Intent} & \multicolumn{1}{c}{Slot} & Overall \\ 
\hline
\multicolumn{7}{c}{\textbf{Autoregressive Models}} \\ 
\hline
Joint Seq \cite{Multi-domainJoint2016}          & 92.6       & 94.2          & 80.7        & 96.9       & 87.3          & 73.2   \\
Attention-BiRNN \cite{attention-basedRnn2016}   & 91.1       & 94.2          & 78.9        & 96.7       & 87.8          & 74.1   \\
Slot-Gated \cite{Slot-gated2018}    & 93.6       & 94.8          & 82.2        & 97.0       & 88.8          & 75.5   \\
Bi-model\cite{birnn2018}          & 96.4         & 95.5       & 85.7          & 97.2         & 93.5       & 83.8          \\
SF-ID \cite{bi-dictional2019}       & 97.8       & 95.8          & 86.8        & 97.4       & 92.2          & 80.6   \\
Stack-Propagation \cite{qin2019stack}           & 96.9       & 95.9          & 86.5        & 98.0       & 94.2          & 86.9   \\
AG-IF \cite{qin2020agif} & 97.1      & 95.9     & 87.0       & 98.0       & 94.3          & 87.3   \\
\hline
\multicolumn{7}{c}{\textbf{Non-autoregressive Models}} \\ 
\hline
SlotRefine \cite{wu2020slotrefine}  & 97.1       & 96.0          & 86.9        & 97.4       & 93.5          & 84.4   \\ 
GL-GIN \cite{qin2021glgin} & 95.6 & 95.5 & 85.5    & 97.4 & 93.3 & 84.7\\ 
Basic model  & 96.8      & 95.2      & 85.6      & 96.1      & 92.8      & 82.1 \\
LRT & \ \textbf{98.2}$\uparrow$ & \ \textbf{96.1}$\uparrow$ & \ \textbf{87.2}$\uparrow$ & \ \textbf{98.4}$\uparrow$ & \ \textbf{94.8}$\uparrow$ & \ \textbf{88.4}$\uparrow$ \\
\hline
\end{tabular}
}
\caption{SLU performance on ATIS and SNIPS datasets.
The numbers with $\uparrow$ indicate that the improvement of our model over all baselines is statistically significant with $p < 0.05$ under t-test.}
\label{Table:res0}
\end{table*}
\paragraph{\textbf{SLU Performance:}}
The experiment results on ATIS and SNIPS datasets are shown in Table \ref{Table:res0}.
The results show that our model significantly outperforms all the baselines and achieves the best performance in all three metrics.
Compared with the prior non-autoregressive model SlotRefine, our model enhances the performance by 1.1\% (ID), 0.1\% (SF), and 0.3\% (Overall) on ATIS and 1.0\% (ID), 1.3\% (SF), and 4.0\% (Overall) on SNIPS. Compared with the SOTA baseline AG-IF, LRT also achieve improvement by 1.1\% (ID), 0.2\% (SF), and 0.2\% (Overall) on ATIS and 0.4\% (ID), 0.5\% (SF), and 1.1\% (Overall) on SNIPS.
This indicates the effectiveness of our model.

Notably, without SLG and LRM, our basic model performs worse than both SlotRefine and AG-IF, but LRT outperforms both of them with SLG and LRM. We attribute this enhancement to the fact that our SLG task effectively obtains the sequential dependency information, and LRM directly takes the explicit result information into consideration, which grasps the relationship between the intent and slots. We will conduct experiments for the ablation study in section \ref{section: Ablation Study} to further verify this idea.

\paragraph{\textbf{Speed Up:}}
We also conduct experiments to test the latency of SLU models on single-turn SLU tasks in the same environment as we do on multi-turn SLU tasks.
From Table \ref{Table: res_speed-single}, we can obviously find that our model achieves significant speedup against the autoregressive model AG-IF ($\times$14.04 on ATIS; $\times$13.26 on SNIPS) and Stack-Propagation ($\times$10.61 on ATIS; $\times$10.19 on SNIPS) because all slot labels are calculated simultaneously in our non-autoregressive method.

\begin{table}[h]
\centering
\resizebox{0.6\columnwidth}{!}{
\begin{tabular}{l|cc|cc}
\hline
\multirow{2}{*}{Model} & \multicolumn{2}{c|}{\textbf{ATIS}}          & \multicolumn{2}{c}{\textbf{SNIPS}}         \\ \cline{2-5} 
                       & \multicolumn{1}{c}{Latency } & Speedup & \multicolumn{1}{c}{Latency } & Speedup \\ 
\hline
AG-IF                  & 123.81ms   & 1.00$\times$  & 100.23ms  & 1.00$\times$  \\
Stack-Propagation      & 93.78ms   & 1.32$\times$  & 77.04ms  & 1.30$\times$   \\
SlotRefine             & 21.76ms    & 5.69$\times$  & 22.14ms   &4.53$\times$   \\ 
GL-GIN                 & 12.72ms    & 9.73$\times$  & 9.89ms   &10.13$\times$   \\
\hline
Basic model    & 8.71ms   & 14.21$\times$  & 7.44ms  & 13.47$\times$ \\
LRT    & 8.82ms   & 14.04$\times$  & 7.56ms  & 13.26$\times$  \\
\hline
\end{tabular}
}
\caption{Latency of SLU models on single-turn SLU tasks. ``Latency'' is the average inference time without minibatching. “Speedup” is compared against the existing SOTA model AG-IF \cite{qin2020agif}.}
\label{Table: res_speed-single}
\end{table}


More importantly, compared with the existing non-autoregressive model, our model also achieves speedup against SlotRefine ($\times$2.47 on ATIS; $\times$2.93 on SNIPS) and Gl-GIN ($\times$1.44 on ATIS; $\times$1.31 on SNIPS). The reason for this phenomenon is that SlotRefine needs to run the whole model twice, and GL-GIN contains a LSTM-based Encoder in its structure. As a comparison, our model is lightweight and obtains the final SLU results with a one-period prediction. 

From Table \ref{Table: res_speed-single}, we can also find that utilizing LRM consumes only 3\% extra inference time compared to without LRM. 
This indicates that LRM is a lightweight approach, which generates negligible time cost. 

Above all, our proposed model significantly improves performance while substantially accelerate the inference speed against the existing autoregressive models on both single-turn SLU tasks and multi-turn SLU task. 

\subsection{Error Analysis}
In this section, we will analysis the error caused by uncoordinated slots.

\subsubsection{Error Analysis on Multi-turn SLU Tasks}
\ 
\newline
We first analyze this kind of error in multi-turn SLU tasks.
To analyze the error in SF tasks in detail, we show the statistics of slot error (i.e., incorrect slots) on the validation set of KVRET. Specifically, we define the errors caused by the uncoordinated slot problem as "Unc. error". The uncoordinated slots includes two cases, i.e., correct `\emph{B-tag}' followed wrong `\emph{I-tag}' and wrong `\emph{B-tag}' following correct `\emph{I-tag}'. We define the first case as "BI error" and the second case as "IB error.", respectively.

\begin{table}[h]
\centering
\resizebox{0.7\columnwidth}{!}{
\begin{tabular}{l|ccccc}
\hline
Model  & SF F1 score &slot error   & Unc. error &BI error &IB error\\
\hline
Basic model     & 77.6   & 492    & 58 (12.0\%)  & 39    & 19  \\ 
Basic model + CAT-All & 78.0    & 470   & 67 (14.3\%)  & 54 & 13   \\ 
LRT  & 78.3 & 424   & 28 (6.6\%)  & 17  & 11   \\
LRT+CAT-All & 77.8  & 436   & 38 (8.7\%)  & 27  & 11   \\
\hline
Basic model + SHA    & 95.1   & 70   & 18 (25.7\%)  & 17   & 1  \\
Basic model + SHA-P  & 94.9   & 73   & 15 (20.6\%)  & 14   & 1  \\
\modelname{}   & 96.1   & 67   & 9 (13.4\%)   & 9    & 0  \\
SHA-P-LRT & 95.9   & 68   & 8 (11.8\%)   & 7    & 1  \\
\hline
\end{tabular}
}
\caption{The statistics of slot error on the validation set of KVRET after training 100 epochs.}
\label{Table: unc-problem-multi}
\end{table}

The statistics of slot error on the validation set of KVRET is shown in Table \ref{Table: unc-problem-multi}. From the Table, we can find that our model reduces the uncoordinated slot, including both BI and IB errors. Compared with the Basic model, our model reduces 49 (84.5\%) uncoordinated slots in total, where 76.9\% (30/39) BI error and 100\% (19/19) IB error are solved by our model, respectively. This phenomenon indicates that our model effectively solves the uncoordinated slot problem in multi-turn SLU tasks.

As the Basic model lacks historical information, comparing the Basic model with our SHA-LRT is somehow unfair. Thus, we further count the uncoordinated slots for LRT to make a more direct comparison. Compared with the Basic model, our LRT solves 56.4\% (22/39) BI error and 42.1\% (8/19) IB error and reduces 30 (51.7\%) uncoordinated slots. Meanwhile, we also combine the Basic model with our SHA module and compare it with our SHA-LRT. Even both have the SHA module, our SHA-LRT still reduces about 50\% more Unc. errors than the Basic model.

Moreover, as shown in Table \ref{Table: unc-problem-multi}, concatenating all historical utterances even exacerbates the uncoordinated slot problem. The reason is that the overly long utterance sequence makes it more challenging to obtain the sequence dependency among tokens. 

Note that, for the Basic model, the uncoordinated problem seems not very serious, as Unc. errors account for only 12\% of all slot errors. However, this is mainly because the uncoordinated slot problem requires at least one B-tag or I-tag to be predicted correctly. When the model is not effective enough, it predicts a large number of slot labels which both B-tag and I-tag of it are incorrect. We do not count this kind of slot as the uncoordinated slot, even though the B-tag and I-tag are different, which is `uncoordinated' in a sense.

\subsubsection{Error Analysis on Single-turn SLU Tasks}
\ 
\newline
To avoid the influence of dialogue history and make a more accurate analysis of the uncoordinated slot problem, we further analyze this problem in single-turn SLU tasks.

\begin{figure}[h]
\centering
\begin{tikzpicture}[scale = 0.7]
\begin{axis}[
height=6cm,
width=8cm,
xlabel=Training epochs,
ylabel=Uncoordinated Slots,
ymin=0,
ymax=250,
ytick pos=left,
symbolic x coords={1,5,10,15,20,50,100},
ytick={0,50,100,150,200,250},
ymajorgrids=true,
x tick label style={/pgf/number format/1000 sep={}},
]
\addplot coordinates {
(1, 224)
(5,	88)
(10, 74)
(15,67)
(20,52)
(50,38)
(100,28)
};
\addlegendentry{SlotRefine}


\addplot [color=red,mark=square*]
coordinates {
(1, 222)
(5,	 65)
(10, 47)
(15, 42)
(20, 34)
(50, 25)
(100,13)
};
\addlegendentry{LR-Transformer}

\addplot [color=black,mark=star]
coordinates {
(1, 237)
(5,	 184)
(10, 147)
(15, 112)
(20, 98)
(50, 85)
(100,57)
};
\addlegendentry{Transformer}

\end{axis}
\end{tikzpicture}
\caption{The number of uncoordinated slots on the validation set of SNIPS during training.}
\label{fig: un-slots}
\end{figure}
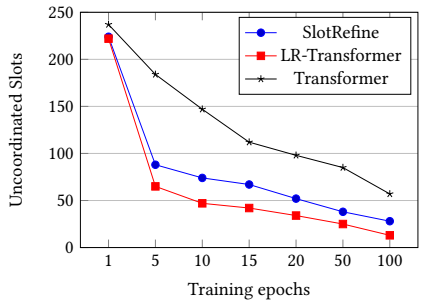

We first visualize the number decrease of uncoordinated slots in the training process. As shown in Fig \ref{fig: un-slots}, the number of uncoordinated slots drops slow and inefficient for the basic model. 
For our LRT, the number of uncoordinated slots drops significantly faster. Ten epochs training of our model is already better than 100 epochs training of the basic model on this problem.
Compared with SlotRefine, our model also drops much faster and achieves better convergence after 5 epochs of training. 

\begin{table}[h]
\centering
\resizebox{0.5\columnwidth}{!}{
\begin{tabular}{l|cccc}
\toprule
Model   &slot error   & Unc. error &BI error &IB error\\
\hline
Basic model         & 169    & 57 (33.7\%)  & 31    & 26   \\ 
SlotRefine             & 127    & 28 (22.0\%)  & 8     & 20    \\ 
LRT      & 117   & 13 (11.1\%)  & 7   & 6  \\
\bottomrule
\end{tabular}
}
\caption{The statistics of slot error on the validation set of SNIPS after training 100 epochs.}
\label{error-analysis}
\end{table}

\begin{figure*}[t]
\centering
\begin{tikzpicture}[node distance = 0cm]

\tikzstyle{Word} = [rectangle,minimum width = 1cm, minimum height = 2cm, text centered, draw = white, fill = white,thick]
\tikzstyle{Back} = [rectangle,minimum width = 15cm, minimum height = 7cm, text centered, draw = black, fill = white,thick]

\tikzstyle{every node}=[scale=0.7]
\node(CENTER)[Back]{};
\node(word1)[Word,below of=CENTER,xshift=-6.7cm,yshift=2cm]{play};
\node(word2)[Word,right of=word1,xshift=1cm,yshift=0cm]{the};
\node(word3)[Word,right of=word2,xshift=1.7cm,yshift=0]{video};
\node(word4)[Word,right of=word3,xshift=2.5cm,yshift=0]{game};
\node(word5)[Word,right of=word4,xshift=2.5cm,yshift=0]{the};
\node(word6)[Word,right of=word5,xshift=2.5cm,yshift=0]{genesis};
\node(word7)[Word,right of=word6,xshift=2.5cm,yshift=0]{machine};
\node(TokenSequence)[below of=word4,xshift=1.15cm,yshift=0.4cm]{\textbf{Token Sequence}:};
\node(slot1)[below of=word1,xshift=0cm,yshift=-1.2cm]{O};
\node(slot2)[right of=slot1,xshift=1cm,yshift=0]{O};
\node(slot3)[right of=slot2,xshift=1.7cm,yshift=0]{B-object\_type};
\node(slot4)[right of=slot3,xshift=2.5cm,yshift=0]{I-object\_type};
\node(slot5)[right of=slot4,xshift=2.5cm,yshift=0]{B-object\_name};
\node(slot6)[right of=slot5,xshift=2.5cm,yshift=0]{I-object\_name};
\node(slot7)[right of=slot6,xshift=2.5cm,yshift=0]{I-object\_name};
\node(Answer)[below of=TokenSequence,xshift=0,yshift=-1.2cm]{\textbf{Corret Slots}:};

\node(wslot1)[below of=slot1,xshift=0cm,yshift=-1.2cm]{O};
\node(wslot2)[right of=wslot1,xshift=1cm,yshift=0]{O};
\node(wslot3)[right of=wslot2,xshift=1.7cm,yshift=0]{B-object\_type};
\node(wslot4)[right of=wslot3,xshift=2.5cm,yshift=0]{{\color{red}I-object\_name}};
\node(wslot5)[right of=wslot4,xshift=2.5cm,yshift=0]{{\color{red}B-object\_type}};
\node(wslot6)[right of=wslot5,xshift=2.5cm,yshift=0]{I-object\_name};
\node(wslot7)[right of=wslot6,xshift=2.5cm,yshift=0]{I-object\_name};
\node(Basic)[below of=Answer,xshift=0,yshift=-1.2cm]{\textbf{Basic Model}:};

\node(srslot1)[below of=wslot1,xshift=0cm,yshift=-1.2cm]{O};
\node(srslot2)[right of=srslot1,xshift=1cm,yshift=0]{O};
\node(srslot3)[right of=srslot2,xshift=1.7cm,yshift=0]{B-object\_type};
\node(srslot4)[right of=srslot3,xshift=2.5cm,yshift=0]{I-object\_type};
\node(srslot5)[right of=srslot4,xshift=2.5cm,yshift=0]{{\color{red}B-object\_type}};
\node(srslot6)[right of=srslot5,xshift=2.5cm,yshift=0]{{\color{red}I-object\_type}};
\node(srslot7)[right of=srslot6,xshift=2.5cm,yshift=0]{{\color{red}I-object\_type}};
\node(SR)[below of=Basic,xshift=0,yshift=-1.1cm]{\textbf{SlotRefine}:};

\node(myslot1)[below of=srslot1,xshift=0cm,yshift=-1.2cm]{O};
\node(myslot2)[right of=myslot1,xshift=1cm,yshift=0]{O};
\node(myslot3)[right of=myslot2,xshift=1.7cm,yshift=0]{B-object\_type};
\node(myslot4)[right of=myslot3,xshift=2.5cm,yshift=0]{I-object\_type};
\node(myslot5)[right of=myslot4,xshift=2.5cm,yshift=0]{B-object\_name};
\node(myslot6)[right of=myslot5,xshift=2.5cm,yshift=0]{I-object\_name};
\node(myslot7)[right of=myslot6,xshift=2.5cm,yshift=0]{I-object\_name};
\node()[below of=SR,xshift=0,yshift=-1.2cm]{\textbf{\modelname{}}:};
\end{tikzpicture}
\caption{Case study for the SF task.}
\label{fig:case-study}
\end{figure*}

To further analyze the error in SF tasks in detail, we show the statistics of slot error on the validation set of SNIPS after training 100 epochs. The statistics show that the uncoordinated slot problem composes a big part of all slot errors. Without any approach to solving this problem, our basic model encounters 57 uncoordinated slots, composing 33.7\% of all slot errors. In these uncoordinated slots, 31 are caused by the "BI error," while the "IB error causes 26." The proportion of the "BI error" and the "IB error" is almost close.

Compared with the basic model, our LRT reduces 44 uncoordinated slots. The proportion of uncoordinated slots in all incorrect slot labels drops from 33.7\% to 11.1\%. Moreover, our LRT efficiently reduces both the BI and IB errors simultaneously. The reducing proportion between the BI error and the IB error of our model is also close. This is mainly because our SLG effectively obtains sequential dependency while LRM considers both B-tag and I-tag.  
As a comparison, SlotRefine corrects most uncoordinated slots caused by the "BI error" but still suffers "IB error." For SlotRefine, the "IB error" proportion is much higher than the "BI error." 

We provide an example for the case study. In Fig \ref{fig:case-study}, we notice that the basic model suffers a serious uncoordinated slot problem, including both two cases, i.e., "BI error" and "IB error," respectively.
SlotRefine solves "BI error" but predicts incorrect slot labels in the "IB error" case due to the error propagation from the wrong B-tag label. Our model solves the problems of both cases and predicts a correct slot label sequence. 

Above all, our LRT indeed remedies the problem of the uncoordinated slots, leading to better performance on SF.

\subsection{Ablation Study}
\label{section: Ablation Study}
In this section, we do the ablation study to verify the effectiveness of SHA, LRM and SLG in detail.

\subsubsection{Ablation Study on Multi-turn SLU Tasks}
\ 
\newline
We first make the ablation study on the multi-turn SLU dataset KVRET.

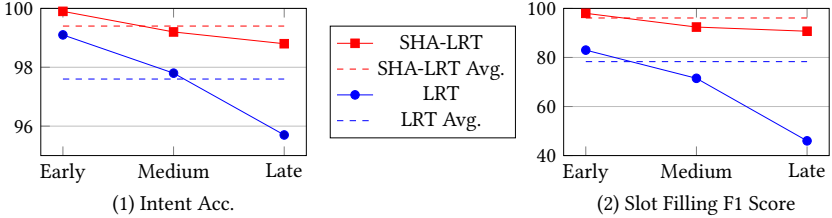
\begin{figure}[h]
\centering
\begin{tikzpicture}[scale = 0.8]
\begin{axis}[
height=4cm,
width=6cm,
xlabel=(1) Intent Acc.,
ymin=95,
ymax=100,
ytick pos=left,
symbolic x coords={Early,Medium,Late},
ymajorgrids=true,
x tick label style={/pgf/number format/1000 sep={}},
]
\addplot 
coordinates {
(Early, 99.1)
(Medium, 97.8)
(Late, 95.7)
};

\addplot [color=blue,dashed]
coordinates {
(Early, 97.6)
(Medium, 97.6)
(Late, 97.6)
};

\addplot [color=red,mark=square*]
coordinates {
(Early, 99.9)
(Medium, 99.2)
(Late, 98.8)
};

\addplot [color=red,dashed]
coordinates {
(Early, 99.4)
(Medium, 99.4)
(Late, 99.4)
};

\end{axis}

\begin{scope}[xshift=8.7cm]
\begin{axis}[
height=4cm,
width=6cm,
xlabel=(2) Slot Filling F1 Score,
ymin=40,
ymax=100,
ytick pos=left,
symbolic x coords={Early,Medium,Late},
ymajorgrids=true,
x tick label style={/pgf/number format/1000 sep={}},
legend style={at={(-0.18,0.5)},anchor=east,scale=0.5} 
]
\addplot [color=red,mark=square*]
coordinates {
(Early, 98.0)
(Medium, 92.4)
(Late, 90.7)
};
\addlegendentry{SHA-LRT}

\addplot [color=red,dashed]
coordinates {
(Early, 96.1)
(Medium, 96.1)
(Late, 96.1)
};
\addlegendentry{SHA-LRT Avg.}

\addplot [color=blue,mark=*]
coordinates {
(Early, 83.0)
(Medium, 71.5)
(Late, 46.0)
};
\addlegendentry{LRT}

\addplot [color=blue,dashed]
coordinates {
(Early, 78.3)
(Medium, 78.3)
(Late, 78.3)
};
\addlegendentry{LRT Avg.}

\end{axis}
\end{scope}

\end{tikzpicture}
\caption{SLU performance of utterance from different part of the full dialogue. The label `Early', `Medium', and `Late' represent the first 1/3 turns, medium 1/3 turns, and the last 1/3 turns, respectively.}
\label{fig:his_effect}
\end{figure}

To begin with, we first analyze how performance varies based on the location of the utterance in the dialogue. As shown in Fig. \ref{fig:his_effect}, the performance of the LRT model drops heavily at the late part of the dialogue in both the ID and SF. This phenomenon is mainly because of two reasons. First, users usually offer more detailed descriptions in the early dialogue, especially for the initial utterance. Moreover, later utterances rely more on salient historical information than utterances in the early dialogue. Compared with LRT, the decline of SHA-LRT is significantly reduced, which initially verifies the significance of our SHA module. We will further verify the effectiveness of our SHA module in the following.

\begin{table*}[h]
\centering
\begin{tabular}{l|ccc}
\hline
\multirow{2}{*}{\textbf{Model}}  & \multicolumn{3}{c}{\textbf{KVRET}}     \\ 
\cline{2-4} & \multicolumn{1}{c}{Intent} & \multicolumn{1}{c}{Slot} & Overall \\
\hline
Basic model  & 97.0      & 77.6     & 75.6 \\
Basic model+RPFSLU          & 96.6  & 75.7  & 74.8 \\
LRT & 97.8 &78.3 &76.4 \\
LRT+RPFSLU & 96.8  &77.1   &75.5 \\
\hline
Basic model + SHA         & 99.0  & 94.0  & 92.7 \\
Basic model + SHA-P       & 98.9  & 93.9  & 92.7 \\
Basic model + SHA\&LRM  & 98.8 & 95.4 & 93.8 \\
Basic model + SHA-P\&LRM & 98.8 & 94.6 & 93.6 \\
Basic model + SHA\&SLG  & 99.1 & 95.5 & 94.0 \\
Basic model + SHA-P\&SLG & 99.3 & 95.1 & 93.1 \\
\hline
\modelname{} (Utterance-Only) & 98.0 & 79.2 & 77.3 \\
\modelname{} (Result-Only) & 98.1 	&88.4 	&85.3 \\
\modelname{} (ResultAttention-Only) & 98.6 & 95.4 & 93.1 \\
\hline
\modelname{} & 99.0 & \textbf{96.1} & \textbf{94.4}\\
SHA-P-LRT & \textbf{99.4} & 95.9 & 94.1 \\
\hline
\end{tabular}
\caption{SLU performance on KVRET datasets with different combinations of modules in \modelname{}.}
\label{Table:ablation-SHA}
\end{table*}

\textbf{Effect of SHA:} From Table. \ref{Table:ablation-SHA}, we find that the RNN-based framework RPFSLU is incompatible with the Transformer-based model. Employing RPFSLU for the Basic model or LRT even reduces prediction accuracy for SLU tasks. This phenomenon further emphasizes the positive meaning of our motivation to design SHA.

In comparison, as mentioned in section \ref{section:Result}, our SHA module significantly enhances the SLU performance for the Transformer-based model. As shown in Table. \ref{Table:ablation-SHA}, compared with only utilize the Basic model, Basic model+SHA enhances the performance by 2.0\%, 16.4\% and 17.1\% (1.9\%, 16.3\%, 17.1\% for SHA-P) on ID,SF and Overall, respectively.
Moreover, compared with LRT, SHA-LRT brings an enhancement of 1.2\%, 17.8\% and 18.0\% (1.6\%, 17.6\%, 17.7\% for SHA-P) on ID,SF and Overall, respectively.
This enhancement is mainly originated from the semantic information contained in the historical utterances and results. SHA effectively incorporates the salient historical information into the model, resulting in a satisfying performance. 

To further detect the origin of this enhancement, we design some variant architectures of SHA as follow:
\begin{itemize}
\item \textbf{\modelname{} (Utterance-Only)}: Remove the History-Result-attention of SHA and direct input $\textbf{H}^u$ into the feed-forward network. In this architecture, only historical utterances will be incorporated into the model.
\item \textbf{\modelname{} (Result-Only)}: Remove the History-Utterance-attention of SHA. Generate the \textbf{Q}, \textbf{K} and \textbf{V} matrix of the History-Result-attention from $\textbf{H}^c$, $\textbf{E}^{r}$ and $\textbf{E}^{r}$, respectively. In this architecture, only historical results will be incorporated into the model.
\item \textbf{\modelname{} (ResultAttention-Only)}: 
Remove the History-Utterance-attention of SHA and direct input $\textbf{H}^c$ into the History-Result-attention. In this architecture, we incorporate historical results into the model, referring to the semantic similarity of historical utterances but ignoring the salient information of the corresponding utterances. 
\end{itemize}

By conducting experiments for these variant models, we find that the effect of only utilizing historical utterances is minimal (\%0.9 on Overall). However, the model has a notable improvement (\%7.6 on Overall) when only utilizing historical results. This result manifests that the enhancement made by SHA is mainly contributed by efficiently exploiting the historical results. Further, ResultAttention-Only enhances the Overall performance of Result-Only by 7.8\%, which verifies the importance of referring to the semantic similarity of historical utterances when doing history-attention. The complete SHA module further enhances ResultAttention-Only by 1.3\% on Overall, indicating historical utterances also contain salient information and need to be concerned.

We then compare the effect of SHA and SHA-P. When LRM and SLG are both utilized, SHA performs better on SF and overall accuracy, and SHA-P performs better on ID. In other cases, SHA performs better on all metrics.

Finally, we verify the compatibility of the SHA module with our other modules. As shown in the third group of Table \ref{Table:ablation-SHA}, we make a permutation for our SLG, LRM, and SHA module. We can first find that both SLG and LRM are able to work non-conflicting with our SHA module and improve the prediction accuracy. Compared with only utilizing the SHA module (i.e., Basic model + SHA/SHA-p), LRM enhances the SF performance by 1.4\%/1.7\% and SLG by 1.5\%/1.2\%. These three modules can also combine as our complete model and further enhance the SLU performance. Therefore, all modules in our \modelname{} have high compatibility. 

\subsubsection{Ablation Study on Single-turn SLU Tasks}
\ 
\newline
Subsequently, we do an ablation study for LRM and SLG. As we have verified the effectiveness of those modules on multi-turn tasks, we make further analysis for LRM and SLG on single-turn datasets in this section. 

\begin{table*}[ht]
\centering
\begin{tabular}{l|ccc|ccc}
\hline
\multirow{2}{*}{\textbf{Model}}  & \multicolumn{3}{c|}{\textbf{ATIS}}    & \multicolumn{3}{c}{\textbf{SNIPS}} \\ 
\cline{2-7} & \multicolumn{1}{c}{Intent} & \multicolumn{1}{c}{Slot} & Overall & \multicolumn{1}{c}{Intent} & \multicolumn{1}{c}{Slot} & Overall \\ 
\hline
Basic model \cite{shaw2018position_relative_attention}   & 96.8      & 94.8      & 85.3      & 96.1      & 92.8      & 82.1 \\
\hline
Basic model + LRM   & $97.7_{0.047}$      & $95.6_{0.123}$       & $86.1_{0.082}$       & $98.0_{0.094}$       & $93.6_{0.125}$       & $84.9_{0.216}$  \\
Basic model + SLG*    & $97.0_{0.094}$       & $95.9_{0.169}$       & $86.2_{0.216}$       & $97.6_{0.124}$       & $94.1_{0.249}$       & $86.3_{0.205}$  \\
Basic model + SLG    & $97.1_{0.047}$      & $95.9_{0.094}$       & $86.3_{0.094}$       & $97.7_{0.047}$      & $94.2_{0.047}$       & $86.5_{0.094}$  \\
LRT* & ${98.1}_{0.047}$  & ${96.1}_{0.163}$  & ${87.1}_{0.141}$  & ${98.7}_{0.081}$  & ${94.6}_{0.169}$  & ${88.2}_{0.169}$  \\
\hline
LRT & $\textbf{98.2}_{0.047}$  & $\textbf{96.1}_{0.047}$ & $\textbf{87.2}_{0.081}$  & $\textbf{98.8}_{0.047}$  & $\textbf{94.8}_{0.094}$  & $\textbf{88.4}_{0.081}$  \\
\hline
\end{tabular}
\caption{Performance (mean and standard deviation of the model repeated 5 times) comparison of each module in our model. Models with * represent not using consistency loss in Eq.\ref{eq_slgloss}, i.e., $\alpha = 0$ in Eq.\ref{eq_slgloss}.}
\label{res_abl}
\end{table*}
\textbf{Effect of LRM:}
By comparing basic model + LRM with the basic model, we find that LRM enhances the performance by 0.9\% (ID), 0.8\% (SF), and 0.8\% (Overall) on ATIS and 1.9\% (ID), 0.8\% (SF), and 2.8\% (Overall) on SNIPS, which shows that utilizing LRM alone still enhances SLU performance on both two tasks. We attribute the improvement to the direct utilization of ID and SF preliminary results. The semantic information from both slot labels and ID labels enhances the performance when predicting.

\begin{figure}[h]
\centering
\begin{tikzpicture}[scale = 0.8]
\begin{axis}[
height=4cm,
width=7cm,
xlabel=Interval of Transformer encoder layer,
ylabel=F1 score,
ymin=92.0,
ymax=95,
ytick pos=left,
symbolic x coords={1-2,2-3,3-4,4-5,5-6},
x tick label style={/pgf/number format/1000 sep={}},
]
\addplot[color=red,mark=*] coordinates {
(1-2,93.6)
(2-3,94.4)
(3-4,93.2)
(4-5,93.5)
(5-6,94.2)
};

\end{axis}
\end{tikzpicture}
\caption{SF performance comparison of using LRM in different Transformer encoder intervals on SNIPS validation set. 1-2 represents the interval of the 1st and the 2nd Transformer encoder layer, and so on.}
\label{fig:LRMpos}
\end{figure}
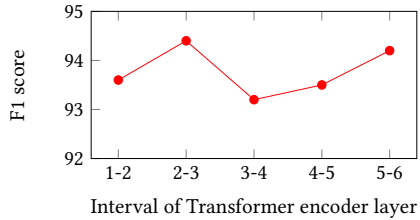

As we introduced in section \ref{sec:LRM}, LRM can be utilized in each two Transformer encoder layer intervals. Motivated by finding the best place to operate LRM, we conduct the experiment to compare SF performance on the validation set of SNIPS when using LRM in different Transformer encoder intervals. The experiment result is shown in Fig. \ref{fig:LRMpos}. 

\begin{figure}[h]
\centering
\begin{tikzpicture}[scale = 0.8]
\begin{axis}[
height=4cm,
width=7cm,
xlabel=Usage count of LRM,
ylabel=F1 score,
ymin=92.0,
ymax=95.0,
ytick pos=left,
symbolic x coords={1,2,3,4,5},
x tick label style={/pgf/number format/1000 sep={}},
]
\addplot[color=red,mark=*] coordinates {
(1, 94.4)
(2,	94.3)
(3, 93.5)
(4,93.2)
(5,92.1)
};

\end{axis}
\end{tikzpicture}
\caption{SF performance comparison of different LRM usage count on SNIPS validation set.}
\label{fig:LRMnum}
\end{figure}
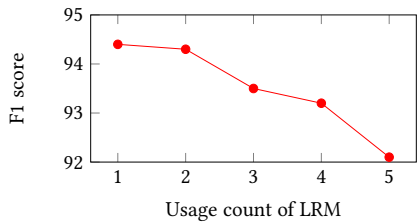

We further conduct experiments to evaluate SF performance for different usage counts of LRM. As shown in Fig. \ref{fig:LRMnum}, the F1 score on the validation set achieves a satisfying level when employing LRM once or twice and drops significantly with increasing usage count.
This phenomenon indicates that employing LRM one time is enough, and overuse LRM will negatively influence SLU prediction. We have introduced the reason in section 3.3, using LRM once in the whole Transformer structure already considers the interaction between ID and SF. Overusing LRM  will make the Transformer layers lose their own feature and negatively affect the final performance due to the utilization of the SLU classifier in LRM.

\textbf{Effect of SLG:}
As shown in Table \ref{res_abl}, compared with the basic model, the basic model + SLG also enhances the performance. Specifically, SLG brings enhancement by 0.3\% for ID, 1.1\% for SF, and 1.0\% for overall on ATIS, while 1.6\% for ID, 1.4\% for SF, and 4.4\% for overall on SNIPS. 
We attribute this improvement to the autoregressive structure of SLG, which brings sequential solid dependency information, making prediction more accurate.

Moreover, comparing LRT with LRT* and LRT+SLG with LRT+SLG*, we find that utilizing consistency loss function in Eq.\ref{eq_slgloss} brings a slight enhancement. More importantly, the consistency loss effectively reduces the standard deviation, especially on SF and overall. This is mainly because utilizing consistency loss can enhance the prediction consistency between the SLU and SLG tasks, making the model more stable. Moreover, the standard deviation on ID tasks is much lower compared with SF. We consider this is because ID contains fewer categories and is easier to be predicted. 

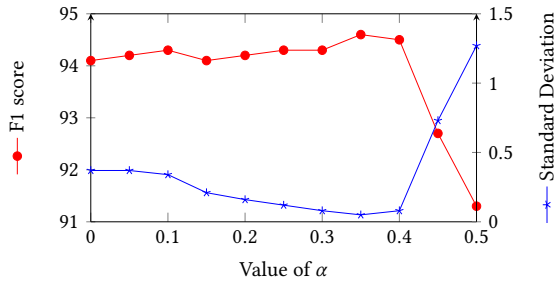
\begin{figure}[h]
\centering
\begin{tikzpicture}[scale = 0.8]
\begin{axis}[
height=5cm,
width=8cm,
xlabel=Value of $\alpha$,
ylabel=\ref{y1} F1 score,
axis y line=left, 
xmin=0,xmax=0.5,
ymin=91.0,ymax=95.0,
ytick pos=left,
xtick = {0,0.1,0.2,0.3,0.4,0.5}
]
\addplot[color=red,mark=*] coordinates {
(0, 94.1)
(0.05, 94.2)
(0.1, 94.3)
(0.15, 94.1)
(0.2, 94.2)
(0.25, 94.3)
(0.3, 94.3)
(0.35, 94.6)
(0.4, 94.5)
(0.45, 92.7)
(0.5, 91.3)
};\label{y1}
\end{axis}

\begin{axis}[
height=5cm,
width=8cm,
axis x line=none,
axis y line=right, 
ylabel=\ref{y2} Standard Deviation,
ymin=0,ymax=1.5,
ytick pos=right,
xmin=0,xmax=0.5,
xtick = {0,0.05,0.1,0.15,0.2,0.25,0.3,0.35,0.4,0.45,0.5}
]
\addplot[color=blue,mark=star] coordinates {
(0, 0.37)
(0.05, 0.37)
(0.1, 0.34)
(0.15, 0.21)
(0.2, 0.16)
(0.25, 0.12)
(0.3, 0.08)
(0.35, 0.05)
(0.4, 0.08)
(0.45, 0.73)
(0.5, 1.27)
};\label{y2}
\end{axis}

\end{tikzpicture}
\caption{Impact of $\alpha$ on the SF performance on the SNIPS validation set.}
\label{fig:alpha}
\end{figure}

We further conduct experiments to study the impact of $\alpha$ on the SF performance. In this experiment, we keep the other settings and change the $\alpha$ from 0 to 0.5. According to the results shown in Fig \ref{fig:alpha}, we find that the F1 scores change little when $\alpha$ increases from 0 to 0.35, but the stand deviation drops continuously. Then, when $\alpha$ is larger than 0.4, both the F1 score and the stand deviation worsen rapidly. We consider the large weight of $\alpha$ makes the wrong prediction in SLU excessively affect the SLG task and bring negative effect since the predicted labels of SLU are not the same as the ground-truth labels.

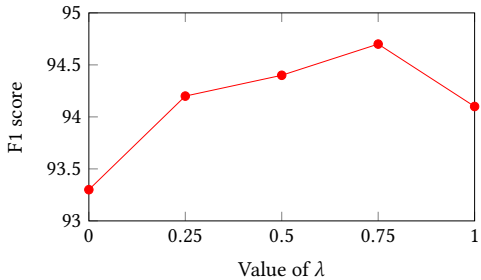
\begin{figure}[h]
\centering
\begin{tikzpicture}[scale = 0.8]
\begin{axis}[
height=5cm,
width=8cm,
xlabel=Value of $\lambda$,
ylabel=F1 score,
xmin=0,
xmax=1,
ymin=93.0,
ymax=95.0,
ytick pos=left,
xtick={0,0.25,0.5,0.75,1},
]
\addplot[color=red,mark=*] coordinates {
(0, 93.3)
(0.25, 94.2)
(0.5, 94.4)
(0.75, 94.7)
(1.0, 94.1)
};

\end{axis}
\end{tikzpicture}
\caption{Impact of $\lambda$ on the SF performance on the SNIPS validation set.}
\label{fig:lambda}
\end{figure}

Similar to $\alpha$, we also conduct experiments to study the impact of $\lambda$ on the SF performance. As shown in Fig. \ref{fig:lambda}, the F1 score enhances when $\lambda$ changes from 0 to 0.25, which indicates the effect of our SLG module. Then, the F1 score continues to rise until $\lambda$ reaches 0.75 and begins to drop with the increase of $\lambda$. This observation indicates that we should make a suitable balance between the SLU and SLG in our final framework.

\subsection{Effect of Pretraining}

In this section, we conduct experiments to evaluate our model's ability to combine the pre-trained model. 

Recently, some work builds their models based on the large-scale pre-trained models, e.g., BERT \cite{BERT}, RoBERTa \cite{roberta}, XLNet \cite{xlnet}, etc., which utilized billions of external corpus and tremendous model parameters. Since the number of parameters in these models is much more than ours, it is unfair to compare the performance of our model with them directly. Thus, we also combine our approaches with pre-trained models to highlight the effectiveness of \modelname{}.

We first conduct experiments on the multi-turn SLU dataset KVRET. 
In practice, we combine pre-trained models with \modelname{} and fine-tune the complete model on the SLU dataset. 
More specifically, we employ pre-trained models to obtain hidden states of the utterance. Then we replace the embedding sequence with the pre-trained hidden states as the input of our SHA module to predict SLU results.

For baselines, we first select BERT-SLU \cite{chen2019bert}, then we replace BERT in BERT-SLU with RoBERTa and XLNet to construct RoBERTa-SLU and XLNet-SLU, respectively. 

\begin{table*}[h]
\centering
\begin{tabular}{l|ccc}
\hline
\multirow{2}{*}{\textbf{Model}}  & \multicolumn{3}{c}{\textbf{KVRET}}     \\ 
\cline{2-4} & \multicolumn{1}{c}{Intent} & \multicolumn{1}{c}{Slot} & Overall \\
\hline
\modelname{} & 99.0 & 96.1 & 94.4\\
SHA-P-LRT & 99.4 & 95.9 & 94.1 \\
\hline
BERT-SLU & 97.8 & 78.6 & 77.0 \\
BERT-SLU + CAT-All & 98.2 & 79.8 & 77.3 \\
BERT+\modelname{} & 98.9 & 98.8 & 96.9 \\
BERT+SHA-P-LRT & 98.9 & 98.3 & 96.8 \\
\hline
RoBERTa-SLU & 97.9 	&80.2 	&77.7 \\
RoBERTa-SLU + CAT-All & 97.8 & 80.3 & 77.4 \\
RoBERTa+\modelname{} & 99.0 & 99.3 & 98.2\\
RoBERTa+SHA-P-LRT & 98.9 & 99.3 & 97.9  \\
\hline
XLNet-SLU & 97.8 & 79.0 & 77.3 \\
XLNet-SLU + CAT-All & 97.3 & 79.1 & 76.7 \\
XLNet+\modelname{} & \textbf{99.9}$\uparrow$ & \textbf{99.6}$\uparrow$ & \textbf{99.5}$\uparrow$ \\
XLNet+SHA-P-LRT & 99.8 & 99.6 & 99.0 \\
\hline
\end{tabular}
\caption{SLU performance for pre-trained models on KVRET datasets.
The numbers with $\uparrow$ indicate that the improvement of our model over all baselines is statistically significant with $p < 0.05$ under t-test.}
\label{Table: res-pretrain-multi}
\end{table*}

The experiment results are shown in Table \ref{Table: res-pretrain-multi}. 
From Table \ref{Table: res-pretrain-multi}, we find that our model is benefited from all three pre-train models. More specifically, the Overall of \modelname{} (SHA/SHA-P) is enhanced by 2.8\%/2.7\% with BERT, 3.8\%/3.5\% with RoBERTa, and 5.4\%/4.9\% with XLNet, respectively. These results intuitively indicate that the pre-trained model further improves the effectiveness of our model, which verifies the compatibility between our model and the pre-trained model.

Besides, as shown in Table \ref{Table: res-pretrain-multi}, the performance of BERT-SLU/RoBERTa-SLU/XLNet-SLU in multi-turn SLU tasks results is unsatisfied due to lacking historical information. Even dealing with dialogue history by CAT-All, the performance of these models is still suffering low accuracy. This phenomenon verifies the effectiveness of SHA once again.

Moreover, to further evaluate the ability of \modelname{} to combine the pre-trained model, we conduct experiments in the single-turn SLU tasks. 
To better use the pre-trained results, we retain the weights of the original pre-trained model and do not change their inner structure. 
Therefore, we only combine our SLG module with the pre-trained model. In practice, we employ pre-trained models as the encoder and utilize the same Transformer decoder as we introduced in section \ref{sec:SLG}. Then we fine-tune the complete model on the SLU dataset. 

\begin{table*}[h]
\centering
\begin{tabular}{l|ccc|ccc}
\hline
\multirow{2}{*}{\textbf{Model}}  & \multicolumn{3}{c|}{\textbf{ATIS}}    & \multicolumn{3}{c}{\textbf{SNIPS}} \\ 
\cline{2-7} & \multicolumn{1}{c}{Intent} & \multicolumn{1}{c}{Slot} & Overall & \multicolumn{1}{c}{Intent} & \multicolumn{1}{c}{Slot} & Overall \\ \hline
\modelname{}   & 98.2 & 96.1 & 87.2      & 98.4 & 94.8 & 88.4\\
\hline
BERT-SLU \cite{chen2019bert}       & 97.5       & 96.1          & 88.2        & 98.6       & 97.0          & 92.8   \\
BERT + Stack-Propagation \cite{qin2019stack}  & 97.5 & 96.1 & 88.6       & 99.0 & 97.0 & 92.9   \\
BERT + SlotRefine  \cite{wu2020slotrefine}  & 97.7 & 96.1 & 88.6         & 99.0 & 97.0 & 92.9   \\
BERT + SLG & \ \textbf{97.9}$\uparrow$ & 96.1 & \ \textbf{88.7}$\uparrow$ & \ \textbf{99.1}$\uparrow$ & \ \textbf{97.2}$\uparrow$ & \ \textbf{93.1}$\uparrow$ \\
\hline
RoBERTa-SLU       & 97.5 & 95.9 & 88.1    & 98.7 & 96.4 & 92.0 \\
RoBERTa + SlotRefine    & 97.6 & 96.0 & 88.2    & 98.9 & 96.4 & 92.1 \\
RoBERTa + SLG   & 97.6 & 96.1 & 88.2      & 99.0 & 96.5 & 92.8 \\
\hline
XLNet-SLU        & 97.5 & 95.1 & 85.2       & 97.1 & 92.9 & 82.7\\
XLNet + SlotRefine    & 97.6 & 95.1 & 85.3    & 97.2 & 92.9 & 82.9\\
XLNet + SLG   & 97.6 & 95.2 & 85.7      & 97.4 & 93.8 & 85.0 \\
\hline
\end{tabular}
\caption{SLU performance of Bert-based models on ATIS and SNIPS datasets. The numbers with $\uparrow$ indicate that the improvement of our model over all baselines is statistically significant with $p < 0.05$ under t-test.}
\label{Table:res_bert}
\end{table*}

As shown in Table \ref{Table:res_bert}, BERT+SLG outperforms all previous BERT-based models on all evaluation metrics. Compared with BERT-SLU, SLG brings an enhancement of 0.8\% (ID), 0.1\% (SF), and 0.5\% (Overall) on ATIS dataset and 0.5\% (ID), 0.1\% (SF), and 0.3\% (Overall) on SNIPS dataset. All of these improvements of our model are statistically significant with $p<0.05$ under t-test. Similarly, RoBERTa+SLG also performs better than all previous RoBERTa-based models. Note that XLNet-based models perform terribly compared with BERT and RoBERTa. This is mainly because the amount of parameters in XLNet is too big, and the model can not be well finetuned with these two datasets. However, SLG still enhances XLNet, and the combined model XLNet+SLG performs better than XLNet+SlotRefine.
The results show that our SLG approach is still useful for pre-trained models, which indicates that our SLG module is portable and effective.

From Table \ref{Table:res_bert}, we further find that the improvement of pre-trained models on the SNIPS dataset is much more obvious than that on the ATIS dataset. Considering that SNIPS contains much more Out-of-Vocabulary (OOV) tokens, we attribute the improvement of BERT to alleviating the OOV problem by the WordPiece encoding \cite{schuster2012japanese}.

\begin{table}[h]
\centering
\begin{tabular}{l|cc|cc}
\hline
\multirow{2}{*}{Model} & \multicolumn{2}{c|}{\textbf{ATIS}}          & \multicolumn{2}{c}{\textbf{SNIPS}}         \\ \cline{2-5} 
                       & \multicolumn{1}{c}{Latency } & Speedup & \multicolumn{1}{c}{Latency } & Speedup \\ 
\hline
BERT+Stack-Propagation      & 220.11ms   & 1.00$\times$  & 225.87ms  & 1.00$\times$   \\
BERT-SLU        & 48.90ms   & 4.50$\times$  & 49.59ms  & 4.55$\times$ \\
BERT+SlotRefine             & 97.81ms    & 2.25$\times$  & 99.19ms   & 2.27$\times$   \\ 
\hline
LRT    & 13.11ms   & -  & 14.01ms  & -  \\
BERT+SLG    & 48.89ms   & 4.50$\times$  & 49.61ms  & 4.55$\times$ \\
\hline
\end{tabular}
\caption{Latency of Bert-based SLU models. ``Latency'' is the average inference time without minibatching. ``Speedup'' is compared against the existing SOTA model BERT+Stack-Propagation.}
\label{Table:res_speed3}
\end{table}
To compare the inference speed with existing SLU models,
we also record the inference time for each SLU model when employing BERT.
As shown in Table \ref{Table:res_speed3}, our BERT+SLG also achieves significant speedup ($\times$4.50 on ATIS; $\times$4.55 on SNIPS) against the autoregressive SOTA model Stack-Propagation. Meanwhile, compared with the existing non-autoregressive model SlotRefine, our model can reduce 50\% inference time, as our model does not need to run BERT two times. 
It is worth noting that our model consumes almost the same time as BERT-SLU \cite{chen2019bert} because our SLG module works only in the training period and wastes no extra time during inference.

Besides, although BERT can bring enhancement, the trade-off for this enhancement is 3.77 times of inference latency when comparing BERT+SLG with \modelname{} in Table \ref{Table:res_speed3}.

To sum up, our proposed model is well combined with the pre-trained models and outperforms all existing models on both multi-turn and single-turn SLU tasks.

\section{Conclusion}
\label{sec: Conclusion}
In this paper, we propose a fast and accurate non-autoregressive model for multi-turn SLU task,i.e., \modelname{}. 
To capture the salient information in the dialogue history, we design the Salient History Attention module, which significantly enhances the multi-turn-SLU performance by taking advantage of historical utterances and historical results with a History-attention approach.
To solve the uncoordinated slot problem, we propose a Layer-Refined Mechanism and the Slot Generation task, respectively.
LRM guides the final prediction via the interaction of intermediate predicted results between Transformer layers. By this process, LRM explicitly introduces the correlation between ID and SF into the model with a little cost of time. 
SLG addresses the sequential dependency information and effectively enhances the performance of SF through multi-task learning with no extra inference time. 
Experiments on KVRET datasets indicate that our model outperforms other previous multi-turn SLU methods by a large margin on both SLU performance and the inference speed. Moreover, we conduct experiments on two public single-turn SLU datasets and further find that our model is also effective on single-turn SLU tasks, which manifests the robustness of our model.

\section{Acknowledgements}
This work is supported by Guangdong Key Lab of AI and Multi-modal Data Processing, United International College (UIC), Zhuhai, Project No. 2020KSYS007, Chinese National Research Fund (NSFC) Project No. 61872239; The Institute of Artificial Intelligence and Future Networks funded by Beijing Normal University (Zhuhai) Guangdong, China; Zhuhai Science-Tech Innovation Bureau, Nos. ZH22017001210119PWC and 28712217900001.

\bibliographystyle{ACM-Reference-Format}
\balance
\bibliography{main.bib}

\end{document}